\documentclass[10pt,twocolumn,letterpaper]{article}

\usepackage{iccv}
\usepackage{times}
\usepackage{epsfig}
\usepackage{graphicx}
\usepackage{amsmath}
\usepackage{amssymb}
\usepackage{algorithm}
\usepackage{algorithmicx}
\usepackage[noend]{algpseudocode}

\usepackage{booktabs}

\usepackage{caption}
\usepackage{subcaption}
\usepackage[dvipsnames]{xcolor}

\usepackage{pifont}
\usepackage{enumitem}
\usepackage{xfrac} 
\usepackage{makecell}
\setlist[itemize]{leftmargin=*}

\usepackage{multirow}
\usepackage{color}
\usepackage{xcolor}
\usepackage{colortbl}
\definecolor{color3}{rgb}{0.95,0.95,0.95}

\usepackage{tipa}

\usepackage[title]{appendix}
\usepackage[accsupp]{axessibility}

\usepackage[breaklinks=true,bookmarks=false]{hyperref}

\iccvfinalcopy 

\def\iccvPaperID{6537} 

\ificcvfinal\fi

\begin{document}

\title{FreeDoM: Training-Free Energy-Guided Conditional Diffusion Model}

\author{Jiwen Yu$^1$\qquad\quad Yinhuai Wang$^1$\qquad\quad Chen Zhao$^2$ \quad\quad Bernard Ghanem$^2$ \quad\quad  Jian Zhang$^{1}$\\
$^1$ Peking University Shenzhen Graduate School \qquad $^2$ KAUST\\
\url{https://github.com/vvictoryuki/FreeDoM}
}

\ificcvfinal\thispagestyle{empty}\fi


\twocolumn[{
\renewcommand\twocolumn[1][]{#1}
\maketitle
\centering
\vspace{-0.4cm}
\includegraphics[width=\textwidth]{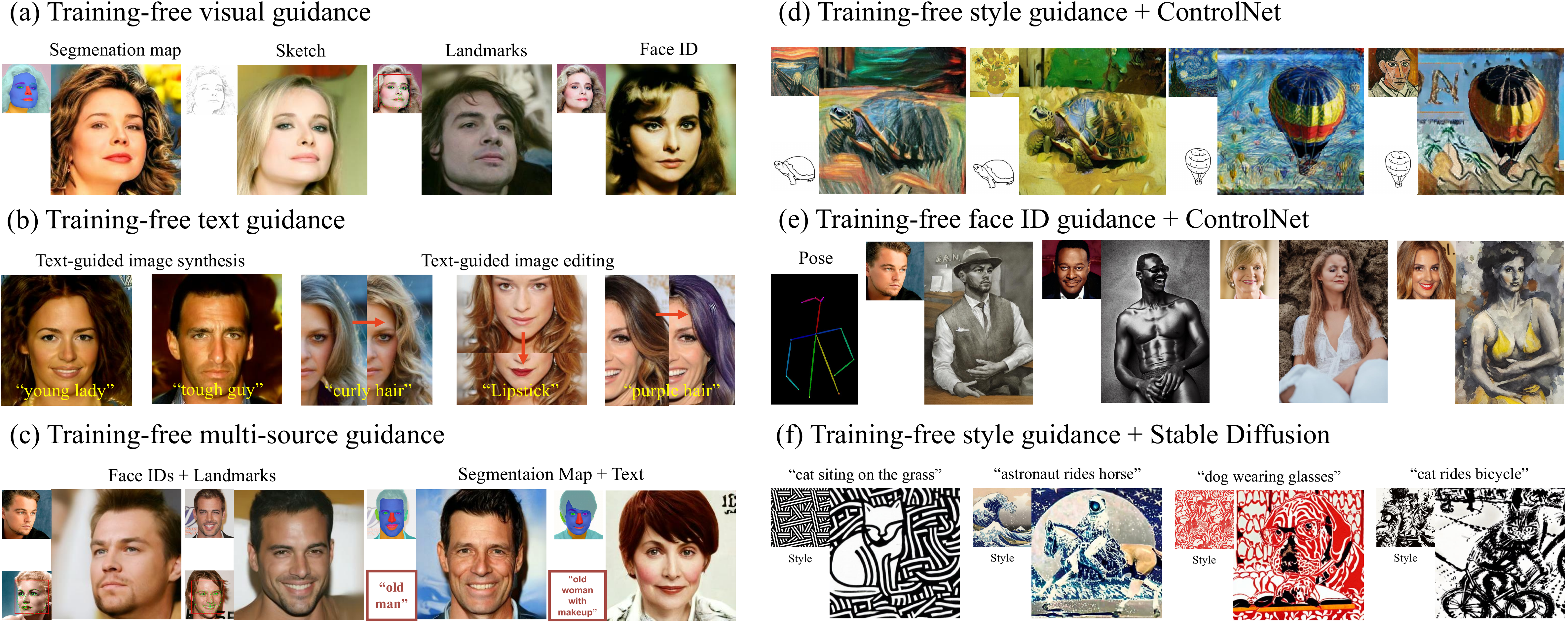}
\vspace{-0.5cm}
\captionsetup{type=figure}
\caption{\textbf{FreeDoM controls the generation process of diffusion models in a training-free way.} Here, we demonstrate some 
results of the applications FreeDoM supports. Part (a)-(c) show various face editing applications with training-free guidance.
(a) We use the segmentation map, sketch, landmarks, and face ID as conditions to guide the generation process of an unconditional diffusion model; 
(b) We use CLIP~\cite{radford2021learning} based text guidance to control image synthesis and editing. For editing, we use the segmentation masks to limit the editing areas (see Fig.~\ref{fig:trick_segmentation} for details);
(c) We combine different conditions to control the generation process.
Part (d)-(f) show that training-free guidance can work with other training-required conditional diffusion models, like Stable Diffusion~\cite{rombach2022high} and ControlNet~\cite{zhang2023adding}, to achieve a more sophisticated control mechanism. 
The conditions of scribbles in (d), human poses in (e), and prompt texts in (f) are controlled by the training-required interfaces provided by ControlNet and Stable Diffusion. Training-free energy functions control the conditions of face IDs from the reference images in (e) and style images in (d) and (f).
\textbf{Zoom in for best view.} 
}
\label{fig:overview}

\vspace{0.9cm}
}]

\begin{abstract}
Recently, conditional diffusion models have gained popularity in numerous applications due to their exceptional generation ability. 
However, many existing methods are training-required. They need to train a time-dependent classifier or a condition-dependent score estimator, which increases the cost of constructing conditional diffusion models and is inconvenient to transfer across different conditions.
Some current works aim to overcome this limitation by proposing training-free solutions, but most can only be applied to a specific category of tasks and not to more general conditions.
In this work, we propose a training-\textbf{Free} conditional \textbf{D}iffusi\textbf{o}n \textbf{M}odel (\textbf{Free}\textbf{Do}\textbf{M}) used for various conditions. 
Specifically, we leverage off-the-shelf pre-trained networks, such as a face detection model, to construct time-independent energy functions, which guide the generation process without requiring training.
Furthermore, because the construction of the energy function is very flexible and adaptable to various conditions, our proposed FreeDoM has a broader range of applications than existing training-free methods. 
FreeDoM is advantageous in its simplicity, effectiveness, and low cost. Experiments demonstrate that FreeDoM is effective for various conditions and suitable for diffusion models of diverse data domains, including image and latent code domains.
\end{abstract}
\vspace{-1cm}

\section{Introduction}
Recently, diffusion models have been demonstrated to outperform previous state-of-the-art generative models~\cite{dhariwal2021diffusion}, such as GANs \cite{gan,mirza2014conditional,brocklarge}. The impressive generative power of diffusion models~\cite{ho2020denoising, song2019generative, song2021scorebased} has motivated researchers to apply diffusion models to various downstream tasks. Conditional generation is one of the most popular focus areas. Conditional diffusion models (CDMs) with diverse conditions have been proposed, such as text~\cite{Kim_2022_CVPR, avrahami2022blended, gu2021vector, liu2019more, rombach2022high, sheynin2023knndiffusion, ramesh2022hierarchical, saharia2022photorealistic, liu2022compositional, nichol2021glide}, class labels~\cite{dhariwal2021diffusion}, degraded images~\cite{Choi_2021_ICCV, chung2023diffusion, Chung_2022_CVPR, chung2022improving, kawar2022denoising, Lugmayr_2022_CVPR, song2023pseudoinverseguided, song2021solving, wang2023zeroshot, saharia2022palette, saharia2022image, whang2022deblurring}, segmentation maps~\cite{mou2023adapter, zhang2023adding}, landmarks~\cite{mou2023adapter, zhang2023adding}, hand-drawn sketches~\cite{mou2023adapter, zhang2023adding}, style images~\cite{mou2023adapter, zhang2023adding}, etc. 
These CDMs can be roughly divided into two categories: training-required or training-free.

A typical type of training-required CDMs trains a time-dependent classifier to guide the noisy image $\mathbf{x}_t$ toward the given condition $\mathbf{c}$ ~\cite{dhariwal2021diffusion, nichol2021glide, zhao2022egsde, liu2019more}. Another branch of training-required CDMs directly trains a new score estimator $s(\mathbf{x}_t, t, \mathbf{c})$ conditioned on $\mathbf{c}$~\cite{mou2023adapter, ho2022classifier, rombach2022high, saharia2022palette, saharia2022image, whang2022deblurring, zhang2023adding, Kim_2022_CVPR, avrahami2022blended, gu2021vector, ramesh2022hierarchical, nichol2021glide}. 
These methods yield impressive performance but are not flexible. Once a new target condition is needed for generation, they have to retrain or finetune the models, which is inconvenient and expensive.

In contrast, training-free CDMs try to solve the same problems without extra training. \cite{ron2022nulltext, feng2023trainingfree, hertz2022prompt} attempt to use the cross-attention control to realize the conditional generation; 
\cite{Choi_2021_ICCV, chung2023diffusion, Chung_2022_CVPR, chung2022improving, kawar2022denoising, Lugmayr_2022_CVPR, song2023pseudoinverseguided, song2021solving, wang2023zeroshot,
wang2023unlimited} directly modify the intermediate results to achieve zero-shot image restoration; \cite{meng2022sdedit} realizes image translation by adjusting the initial noisy images. While these methods are effective in a single application, they are difficult to generalize to a wider range of conditions, e.g., style, face ID, and segmentation masks.

In order to make CDMs support a wide range of conditions 
in a training-free manner, this paper  proposes a training-\textbf{Free} conditional \textbf{D}iffusi\textbf{o}n \textbf{M}odel (\textbf{Free}\textbf{Do}\textbf{M}) with the following two key points. 
\textit{Firstly}, to emphasize generalization, 
we propose a sampling process guided by the energy function~\cite{zhao2022egsde, lecun2006tutorial}, which is very flexible to construct and can be applied to various conditions.
\textit{Secondly}, to make the proposed method training-free, we use off-the-shelf pre-trained time-independent models, which are easily accessible online, to construct the energy function.

Our FreeDoM has the following advantages: (1) \textbf{Simple and effective.} We only insert a derivative step 
of the energy function gradient into the unconditional sampling process of the original diffusion models. Extensive experiments show its 
effective 
controlling capability. (2) \textbf{Low cost and efficient.} The energy functions we construct are time-independent and do not need to be retrained. The diffusion models we choose do not need to be trained on the desired conditions. Thanks to the efficient time-travel strategy we use for large data domains, the number of sampling steps we use is quite small, which speeds up the sampling process while ensuring good generated results. (3) \textbf{Amenable to a wide range of applications.} The conditions our method supports include, but are not limited to, text, segmentation maps, sketches, landmarks, face IDs, style images, etc. In addition, various complex but interesting applications can be realized by combining multiple conditions. (4) \textbf{Supports different types of diffusion models.} Regardless of the considered data domain, such as human face images, images in ImageNet, or latent codes extracted from an image encoder, extensive experiments demonstrate that our method does well on all of them. 

\section{Related Work}
\label{sec:related_work}

\subsection{Training-Required Methods}
The training-required methods can obtain strong control generation ability thanks to supervised learning with data pairs. One of the most prominent applications of these methods is the text-to-image task. The most widely used text-to-image model, Stable Diffusion~\cite{rombach2022high}, generates high-quality images that conform to the text description by inputting a prompt text. Recent works, such as ControlNet~\cite{zhang2023adding} and T2I-Adapter~\cite{mou2023adapter}, have introduced more training-required conditional interfaces to Stable Diffusion, such as edge maps, segmentation maps, depth maps, etc.

Although these training-required methods can achieve satisfactory control results under trained conditions, the cost of training is still a factor to be considered, especially for the scenario that requires more complex control with multiple conditions. The training-required method is not the cheapest or most convenient solution in practical applications.

\subsection{Training-Free Methods}
The training-free methods develop various interesting technologies to realize the training-free condition generation on some tasks exploiting the unique nature of the diffusion model, namely, the iterative denoising process. 
\cite{hertz2022prompt} proposes to inject the target cross attention maps to the source cross attention maps to solve the prompt-to-prompt task without training. The limitation of this method is that a text prompt is needed to anchor the content of the image to be edited in advance.
DDNM~\cite{wang2023zeroshot} proposes to use the Range-Null Space Decomposition to modify the intermediate results to solve the image restoration in a training-free way. It is based on the degradation operators of image restoration tasks and is hard to be adopted in other applications.
SDEdit~\cite{meng2022sdedit} proposes to adjust the initial noisy images to control the generation process, which is useful in stroke-based image synthesis and editing. Its limitation is that the guidance of stroke is not precise and versatile. 

According to the limitations mentioned above, the training-free CDMs for a broad range of applications need to be studied urgently. We have noticed some recent efforts~\cite{parmar2023zero, bansal2023universal} in this area.
Our FreeDoM has a faster generation speed and applies to a broader range of applications.




\section{Preliminaries}
\subsection{Score-based Diffusion Models}
Score-based Diffusion Models (SBDMs)~\cite{song2019generative, song2021scorebased} are a kind of diffusion model based on score theory, which reveals that the essence of diffusion models is to estimate the score function $\nabla_{\mathbf{x}_{t}}\log p(\mathbf{x}_{t})$, where $\mathbf{x}_t$ is noisy data. 
During the sampling process, SBDMs predict $\mathbf{x}_{t-1}$ from $\mathbf{x}_t$ using the estimated score step by step. In our work, we resort to discrete SBDMs with the setting of DDPM~\cite{ho2020denoising} and its sampling formula is:
\begin{equation}
    \mathbf{x}_{t-1}=(1+\frac{1}{2}\beta_t)\mathbf{x}_t+\beta_t\nabla_{\mathbf{x}_{t}}\log p(\mathbf{x}_{t})+\sqrt{\beta_t}\boldsymbol{\epsilon},
    \label{eq:ddpm_sampling}
\end{equation}
where $\boldsymbol{\epsilon}\sim\mathcal{N}(\mathbf{0}, \mathbf{I})$ is randomly sampled Gaussian noise and $\beta_t\in\mathbb{R}$ is a pre-defined parameter.
In actual implementation, the score function will be estimated using a score estimator $s(\mathbf{x}_t, t)$, that is, $s(\mathbf{x}_t, t)\approx\nabla_{\mathbf{x}_{t}}\log p(\mathbf{x}_{t})$.
However, the original diffusion models can only serve as an unconditional generator with randomly synthesized results.

\subsection{Conditional Score Function}
In order to adapt the generative power of the diffusion models to different downstream tasks, conditional diffusion models (CDMs) are needed. SDE~\cite{song2021scorebased} proposed to control the generated results with a given condition $\mathbf{c}$ by modifying the score function as $\nabla_{\mathbf{x}_{t}}\log p(\mathbf{x}_{t}|\mathbf{c})$. Using the Bayesian formula $p(\mathbf{x}_t|\mathbf{c})=\frac{p(\mathbf{c}|\mathbf{x}_t)p(\mathbf{x}_t)}{p(\mathbf{c})}$, we can rewrite the conditional score function as two terms:
\begin{equation}
    \nabla_{\mathbf{x}_t}\log p(\mathbf{x}_t|\mathbf{c}) = \nabla_{\mathbf{x}_t}\log p(\mathbf{x}_t) + \nabla_{\mathbf{x}_t}\log p(\mathbf{c}|\mathbf{x}_t),
    \label{eq:condition_score_theory}
\end{equation}
where the first term $\nabla_{\mathbf{x}_t}\log p(\mathbf{x}_t)$ can be estimated using the pre-trained unconditional score estimator $s(\cdot, t)$ and the second term $\nabla_{\mathbf{x}_t}\log p(\mathbf{c}|\mathbf{x}_t)$ is the critical part of constructing conditional diffusion models. 
We can interpret the second term  $\nabla_{\mathbf{x}_t}\log p(\mathbf{c}|\mathbf{x}_t)$ as a correction gradient, pointing $\mathbf{x}_t$ to a hyperplane in the data space, where all data are compatible with the given condition $\mathbf{c}$.
Classifier-based methods~\cite{dhariwal2021diffusion, nichol2021glide, zhao2022egsde, liu2019more} train a time-dependent classifier to compute this correction gradient for conditional guidance.

\subsection{Energy Diffusion Guidance}
Modeling the correction gradient $\nabla_{\mathbf{x}_t}\log p(\mathbf{c}|\mathbf{x}_t)$ remains an open question. A flexible and straightforward way is resorting to the energy function~\cite{zhao2022egsde, lecun2006tutorial} as follows: 
\begin{equation}
    p(\mathbf{c}|\mathbf{x}_t)=\frac{\exp\{-\lambda\mathcal{E}(\mathbf{c}, \mathbf{x}_t)\}}{\mathnormal{Z}},
\end{equation}
where $\lambda$ denotes the positive temperature coefficient and $\mathnormal{Z}>0$ denotes a normalizing constant, computed as $\mathnormal{Z}=\int_{\mathbf{c}\in\mathcal{C}}\exp\{-\lambda\mathcal{E}(\mathbf{c}, \mathbf{x}_t)\}$ where $\mathcal{C}$ denotes the domain of the given conditions. 
$\mathcal{E}(\mathbf{c}, \mathbf{x}_t)$ is an energy function that measures the compatibility between the condition $\mathbf{c}$ and the noisy image $\mathbf{x}_t$ --- its value will be smaller when $\mathbf{x}_t$ is more compatible with $\mathbf{c}$. If $\mathbf{x}_t$ satisfies the constraint of $\mathbf{c}$ perfectly, the energy value should be zero. Any function satisfying the above property 
can serve as a feasible energy function, with which we just need to adjust the coefficient $\lambda$ to obtain $p(\mathbf{c}|\mathbf{x}_t)$. 

Therefore, the correction gradient $\nabla_{\mathbf{x}_t}\log p(\mathbf{c}|\mathbf{x}_t)$ can be implemented with the following:
\begin{equation}
    \nabla_{\mathbf{x}_t}\log p(\mathbf{c}|\mathbf{x}_t)\propto -\nabla_{\mathbf{x}_t}\mathcal{E}(\mathbf{c}, \mathbf{x}_t),
    \label{eq:score_approx_energy}
\end{equation}
which is referred to as energy guidance. 
With Eq.~(\ref{eq:ddpm_sampling}), Eq.~(\ref{eq:condition_score_theory}), and Eq.~(\ref{eq:score_approx_energy}), we get the conditional sampling:
\begin{equation}
    \mathbf{x}_{t-1}=\mathbf{m}_t - \rho_t\nabla_{\mathbf{x}_t}\mathcal{E}(\mathbf{c}, \mathbf{x}_t),
    \label{eq:energy_guided_sampling}
\end{equation}
where $\mathbf{m}_t = (1+\frac{1}{2}\beta_t)\mathbf{x}_t+\beta_t\nabla_{\mathbf{x}_{t}}\log p(\mathbf{x}_{t})+\sqrt{\beta_t}\boldsymbol{\epsilon}$, and $\rho_t$ is a scale factor, which  can be seen as the learning rate of the correction term.
Eq.~(\ref{eq:energy_guided_sampling}) is a generic formulation of conditional diffusion models, which enables the use of different  energy functions. 

\section{The Proposed FreeDoM Method}
In Sec.~\ref{subsec:approximation}, we approximate the time-dependent energy function using time-independent distance measuring functions, making our method training-free and flexible  for various conditions. 
In Sec.~\ref{subsec:ttt}, we first analyze the reason why the energy guidance fails in a large data domain and then propose an efficient version of the time-travel strategy~\cite{Lugmayr_2022_CVPR, wang2023zeroshot}.
In Sec.~\ref{subsec:construct_energy_functions}, we describe the details of how to construct the energy functions.
In Sec.~\ref{subsec:specific_examples}, we provide specific examples of supported conditions.

\subsection{Approximate Time-Dependent Energy}
\label{subsec:approximation}

We use the energy function to guide the generation due to its flexibility to  construct and suitability to various conditions. 
Existing 
classifier-based 
methods~\cite{dhariwal2021diffusion, nichol2021glide, zhao2022egsde, liu2019more}  choose time-dependent distance measuring functions $\mathcal{D}_{\boldsymbol{\phi}}(\mathbf{c}, \mathbf{x}_t, t)$ to approximate the energy functions as follows: 
\begin{equation}
    \mathcal{E}(\mathbf{c}, \mathbf{x}_t)\approx  \mathcal{D}_{\boldsymbol{\phi}}(\mathbf{c}, \mathbf{x}_t, t),
    \label{eq:e_st}
\end{equation}
where $\boldsymbol{\phi}$ defines the pre-trained parameters. $\mathcal{D}_{\boldsymbol{\phi}}(\mathbf{c}, \mathbf{x}_t, t)$ computes the distance
between the given condition $\mathbf{c}$ and noisy intermediate results $\mathbf{x}_t$. 
The distance measuring functions for noisy data $\mathbf{x}_t$ cannot be directly constructed because it is difficult to find an existing pre-trained network for noisy images. In this case, we have to train (or finetune) a time-dependent network for each type of condition.


Compared with time-dependent networks, time-independent distance measuring functions for clean data $\mathbf{x}_{0}$ are widely available. Many off-the-shelf pre-trained networks such as classification networks, segmentation networks, and face ID encoding networks are open-source and work well on clean images. We denote these distance measuring networks for clean data as $\mathcal{D}_{\boldsymbol{\theta}}(\mathbf{c}, \mathbf{x}_0)$, where $\boldsymbol{\theta}$ denotes their pre-trained parameters.
To use these networks for the energy functions, a straightforward way is to approximate $\mathcal{D}_{\boldsymbol{\phi}}(\mathbf{c}, \mathbf{x}_t, t)$ using $\mathcal{D}_{\boldsymbol{\theta}}(\mathbf{c}, \mathbf{x}_0)$, formulated as:
\begin{equation}
    \mathcal{D}_{\boldsymbol{\phi}}(\mathbf{c}, \mathbf{x}_t, t)\approx\mathbb{E}_{p(\mathbf{x}_0|\mathbf{x}_t)}[\mathcal{D}_{\boldsymbol{\theta}}(\mathbf{c}, \mathbf{x}_0)]. 
    \label{eq:estimate_noisy_s_using_clean_x0t}
\end{equation}
Eq.~(\ref{eq:estimate_noisy_s_using_clean_x0t}) is reasonable because if the distance between the noise image $\mathbf{x}_t$ and the condition $\mathbf{c}$ is small, the clean image $\mathbf{x}_{0}$ corresponding to the noise image $\mathbf{x}_t$ should also have a small distance with the condition $\mathbf{c}$, especially during the late stage of the sampling process when the noise level of $\mathbf{x}_t$ is relatively small.
However, during the sampling process, it is infeasible to get the clean image $\mathbf{x}_0$ corresponding to an intermediate noisy result $\mathbf{x}_t$, so we need to approximate  $\mathbf{x}_0$.
Considering the expectation of  $p(\mathbf{x}_0|\mathbf{x}_t)$ \cite{chung2023diffusion}: 
\begin{equation}
    \mathbf{x}_{0|t}:=\mathbb{E}[\mathbf{x}_0|\mathbf{x}_t]=\frac{1}{\sqrt{\bar{\alpha}_t}}(\mathbf{x}_t+(1-\bar{\alpha}_t)s(\mathbf{x}_t, t)),
    \label{eq:x0t}
\end{equation}
where $\bar{\alpha}_t = \prod_{i=1}^{t} (1 - \beta_i)$ and $s(\cdot, t)$ is the pre-trained score estimator. According to Eq.~(\ref{eq:x0t}), from $\mathbf{x}_t$, we can estimate the clean image 
denoted as $\mathbf{x}_{0|t}$.
Then with Eq.~(\ref{eq:e_st}) and Eq.~(\ref{eq:estimate_noisy_s_using_clean_x0t}), we can approximate the time-dependent energy function of noisy data $\mathbf{x}_t$: 
\begin{equation}
    \mathcal{E}(\mathbf{c}, \mathbf{x}_t)\approx \mathcal{D}_{\boldsymbol{\theta}}(\mathbf{c}, \mathbf{x}_{0|t}).
    \label{eq:energy_approx_tid_smf}
\end{equation}
According to Eq.~(\ref{eq:energy_guided_sampling}) and Eq.~(\ref{eq:energy_approx_tid_smf}),
the approximated sampling process can be written as:
\begin{equation}
    \mathbf{x}_{t-1} = \mathbf{m}_t - \rho_t\nabla_{\mathbf{x}_t}\mathcal{D}_{\boldsymbol{\theta}}(\mathbf{c}, \mathbf{x}_{0|t}(\mathbf{x}_t)),
    \label{eq:approximated_energy_guided_sampling}
\end{equation}
and the detailed algorithm is shown in \textbf{Algo.}~\ref{alg:aio}.

\begin{algorithm}[t]
\vspace{-0.1cm}
\scriptsize
\caption{Sampling Process of our proposed FreeDoM}
\label{alg:aio}
\begin{algorithmic}[1] 
    \Require condition $\mathbf{c}$, unconditional score estimator $s(\cdot, t)$, time-independent distance measuring function $\mathcal{D}_{\boldsymbol{\theta}}(\mathbf{c}, \cdot)$, pre-defined parameters $\beta_t$, $\bar{\alpha}_t$ and learning rate $\rho_t$.
    \State $\mathbf{x}_{T}\sim\mathcal{N}(\mathbf{0},\mathbf{I})$
    \For{$t = T, ..., 1$}
        \State $\boldsymbol{\epsilon}\sim\mathcal{N}(\mathbf{0},\mathbf{I})$ if $t>1$, else $\boldsymbol{\epsilon}=\mathbf{0}$.
        \State $\mathbf{x}_{t-1} = (1+\frac{1}{2}\beta_t)\mathbf{x}_t+\beta_t s(\mathbf{x}_t, t)+\sqrt{\beta_t}\boldsymbol{\epsilon}$
        \State $\mathbf{x}_{0|t}=\frac{1}{\sqrt{\bar{\alpha}_t}}(\mathbf{x}_t+(1-\bar{\alpha}_t)s(\mathbf{x}_t, t))$
        \State $\boldsymbol{g}_t = \nabla_{\mathbf{x}_t}\mathcal{D}_{\boldsymbol{\theta}}(\mathbf{c}, \mathbf{x}_{0|t}(\mathbf{x}_t)))$
        \State $\mathbf{x}_{t-1} = \mathbf{x}_{t-1} - \rho_t \boldsymbol{g}_t$
    \EndFor
    \State \textbf{return} $\mathbf{x}_{0}$
\end{algorithmic}
\end{algorithm}

\subsection{Efficient Time-Travel Strategy}
\label{subsec:ttt}


 \begin{figure}[!tbp]
  \centering
  \vspace{-0.1cm}
  \includegraphics[width=1\linewidth]{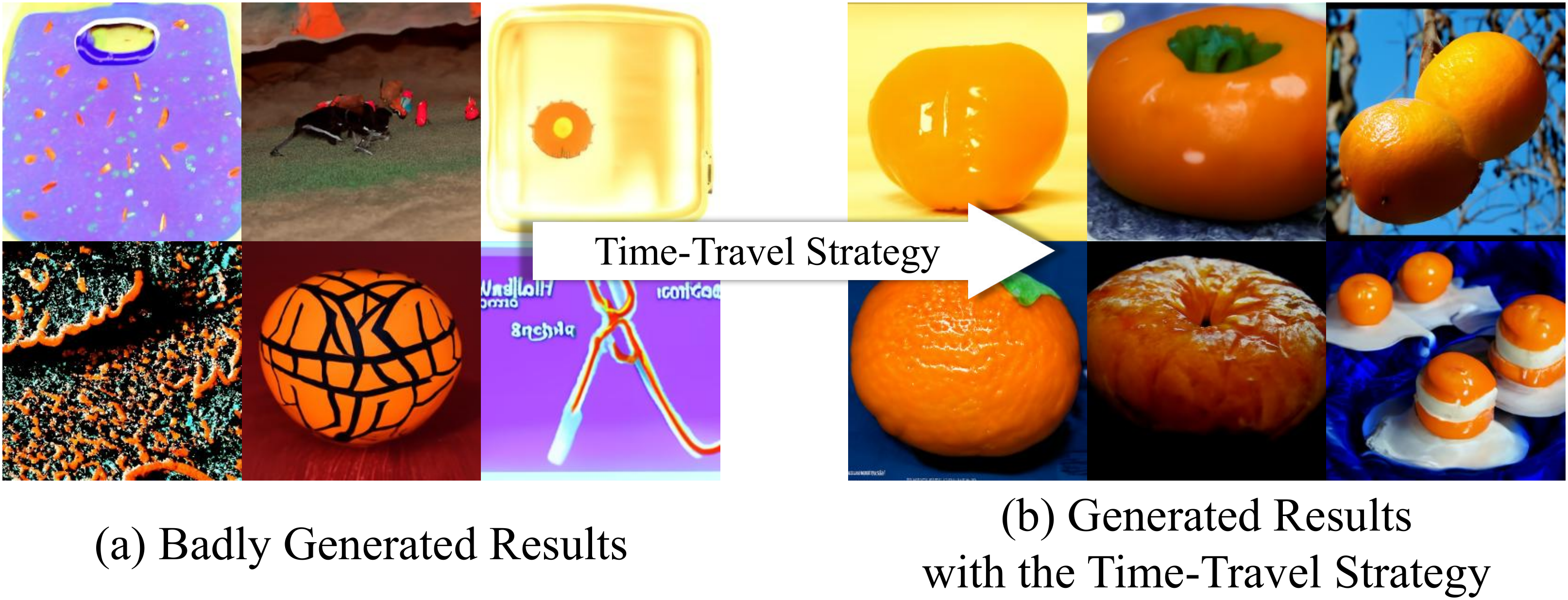}
  \vspace{-0.7cm}
  \caption{\textbf{Comparison of results generated before and after using the time-travel strategy.} The prompt is ``orange''. We can see that the results in (a) do not match the given conditions. After using the time-travel strategy, we get better results in (b).}
  \vspace{-0.1cm}
\label{fig:tt_effective} 
\end{figure}

\begin{figure*}[ht]
  \centering
  \includegraphics[width=1\linewidth]{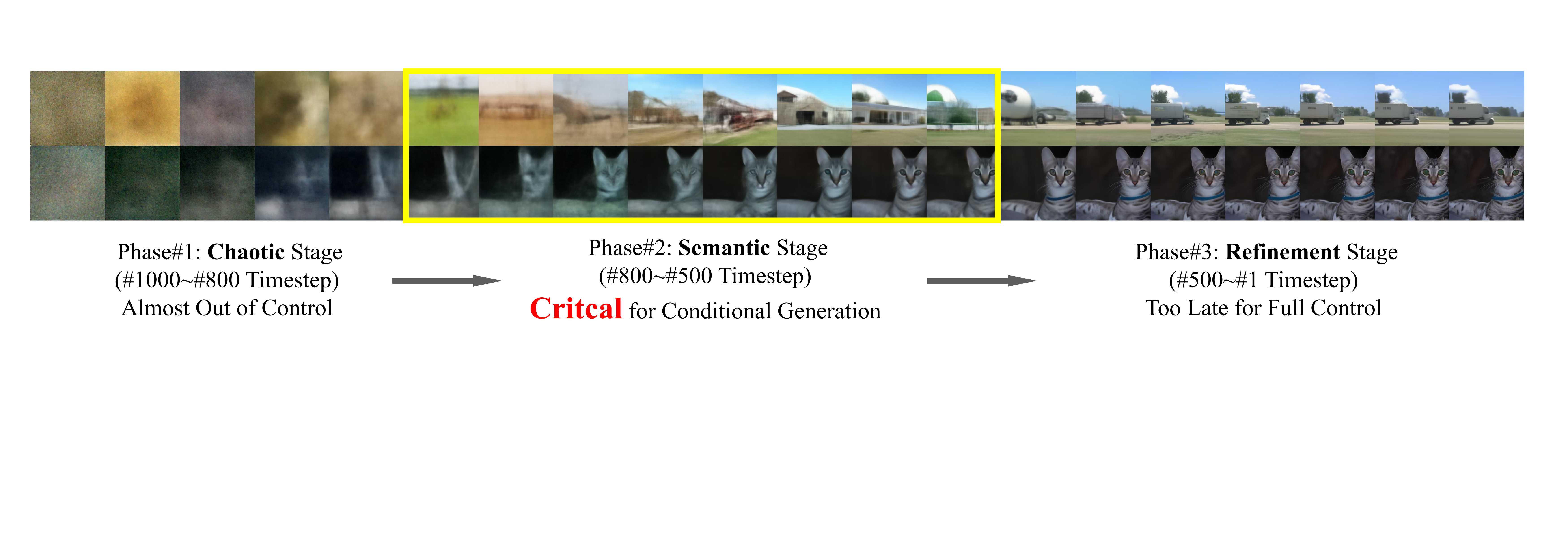}
  \vspace{-0.6cm}
  \caption{\textbf{Demonstration of the importance of different sampling stages.} Most of the semantic content is generated during the semantic stage, so we only employ the time-travel strategy in this stage to achieve an efficient version of FreeDoM. The shown images are $\mathbf{x}_{0|t}$ generated by diffusion models pre-trained on the ImageNet data domain.}
  \vspace{-0.1cm}
\label{fig:ttt2} 
\end{figure*}

In the process of applying \textbf{Algo.}~\ref{alg:aio}, we find that the performance varies significantly on different data domains.  For small data domains such as human faces, \textbf{Algo.}~\ref{alg:aio} can effectively produce results that satisfy the given conditions within 100 DDIM~\cite{song2021denoising} sampling steps. However, for large data domains such as ImageNet, we often get results that are not closely related to the given conditions or even randomly generated results (shown in Fig.~\ref{fig:tt_effective}(a)). 
 We attribute the failure of  \textbf{Algo.}~\ref{alg:aio} on large data domains to poor guidance. 
 The reason for poor guidance is that the direction of unconditional score generated by diffusion models in large data domains has more freedom, making it easier to deviate from the direction of conditional control.
To solve this problem, we adopt the time-travel strategy~\cite{Lugmayr_2022_CVPR, wang2023zeroshot}, which has been empirically shown to inhibit the generation of disharmonious results when solving hard generation tasks.

The time-travel strategy is a technique that takes the current intermediate result $\mathbf{x}_t$ back by $j$ steps to $\mathbf{x}_{t+j}$ and resamples it to the $t$-th timestep again. This strategy inserts more sampling steps into the sampling process and refines the generated results. In our experiments specifically, we go back by $j=1$ step each time and resample. We repeat this resampling process $r_t$ times at the $t$-th timestep. Our experiments demonstrate that the time travel strategy is effective in solving the poor guidance problem (shown in Fig.~\ref{fig:tt_effective}(b)). 
However, the time cost is also expensive because the number of sampling steps is larger, especially considering that each timestep will include the cost of calculating the gradient of the energy function.

Fortunately, we find that the time-travel strategy does not have the same effect in each time step. In fact, using this technique in most time steps will not significantly modify the final result, which means we can use this strategy only in a small portion of  the timesteps, thus significantly reducing the number of additional iteration steps.
In Fig.~\ref{fig:ttt2}, we try to analyze this phenomenon by dividing the sampling process into three stages. In the early stage, i.e., the chaotic stage, the generated result $\mathbf{x}_{0|t}$ is extremely blurred, and the energy guidance is hard to make anything reasonable, so we do not need to employ the time-travel strategy. During the late stage, i.e., the refinement stage, the change in the generated results is minor, so the time-travel strategy is useless.
During the intermediate stage, i.e., the semantic stage, the change in the generated result is significant, so this stage is critical for conditional generation. Based on this observation, we only apply the time-travel strategy in the semantic stage to implement efficient sampling while solving the problem of poor guidance. 
The range of the semantic stage is an experimental choice depending on the specific diffusion models we choose.
The detailed algorithm of our proposed FreeDoM with the efficient time-travel strategy is shown in \textbf{Algo.}~\ref{alg:aio_ttt}, where $r_t=1$ means we do not apply the time-travel strategy in the $t$-th timestep.

\begin{algorithm}[t]
\vspace{-0.1cm}
\scriptsize
\caption{FreeDoM + Efficient Time-Travel Strategy}
\label{alg:aio_ttt}
\begin{algorithmic}[1] 
    \Require condition $\mathbf{c}$, unconditional score estimator $s(\cdot, t)$, time-independent distance measuring function $\mathcal{D}_{\boldsymbol{\theta}}(\mathbf{c}, \cdot)$, pre-defined parameters $\beta_t$, $\bar{\alpha}_t$, learning rate $\rho_t$, and the repeat times of time travel of each step $\{r_1, \cdots, r_T\}$.
    \State $\mathbf{x}_{T}\sim\mathcal{N}(\mathbf{0},\mathbf{I})$
    \For{$t = T, ..., 1$}
        \For{$i = r_t, ..., 1$}
            \State $\boldsymbol{\epsilon}_1\sim\mathcal{N}(\mathbf{0},\mathbf{I})$ if $t>1$, else $\boldsymbol{\epsilon}_1=\mathbf{0}$.
            \State $\mathbf{x}_{t-1} = (1+\frac{1}{2}\beta_t)\mathbf{x}_t+\beta_t s(\mathbf{x}_t, t)+\sqrt{\beta_t}\boldsymbol{\epsilon}_1$
            \State $\mathbf{x}_{0|t}=\frac{1}{\sqrt{\bar{\alpha}_t}}(\mathbf{x}_t+(1-\bar{\alpha}_t)s(\mathbf{x}_t, t))$
            \State $\boldsymbol{g}_t = \nabla_{\mathbf{x}_t}\mathcal{D}_{\boldsymbol{\theta}}(\mathbf{c}, \mathbf{x}_{0|t}(\mathbf{x}_t)))$
            \State $\mathbf{x}_{t-1} = \mathbf{x}_{t-1} - \rho_t \boldsymbol{g}_t$
            \If{$i > 1$}
                \State $\boldsymbol{\epsilon}_2\sim\mathcal{N}(\mathbf{0},\mathbf{I})$
                \State $\mathbf{x}_{t} = \sqrt{1 - \beta_t}\mathbf{x}_{t-1} + \sqrt{\beta_t}\boldsymbol{\epsilon}_2$
            \EndIf
        \EndFor
    \EndFor
    \State \textbf{return} $\mathbf{x}_{0}$
\end{algorithmic}
\end{algorithm}

\subsection{Construction of the Energy Function}
\label{subsec:construct_energy_functions}
\vspace{3pt}
\noindent \ding{113}~\textbf{Single Condition Guidance.}
To incorporate in specific applications, we use the distance measuring function conforming to the following structure to construct the energy function:
\begin{equation}
    \mathcal{E}(\mathbf{c}, \mathbf{x}_t)\approx\mathcal{D}_{\boldsymbol{\theta}}(\mathbf{c}, \mathbf{x}_{0|t})=Dist(\mathcal{P}_{\boldsymbol{\theta}_1}(\mathbf{c}), \mathcal{P}_{\boldsymbol{\theta}_2}(\mathbf{x}_{0|t})),
    \label{eq:single_condition}
\end{equation}
where $Dist(\cdot)$ denotes the distance measuring methods like Euclidean distance, and $\boldsymbol{\theta} = \{\boldsymbol{\theta}_1, \boldsymbol{\theta}_2\}$. $\mathcal{P}_{\boldsymbol{\theta}_1}(\cdot)$ and $\mathcal{P}_{\boldsymbol{\theta}_2}(\cdot)$ project the condition and image to the same space for distance measurement. These projection networks can be off-the-shelf pre-trained classification networks, segmentation networks, etc. In most cases, we  only need one network to project the clean image $\mathbf{x}_{0|t}$ to the condition space. In the cases with reference images $\mathbf{x}_{ref}$, we also  only need one feature encoder to project the reference image $\mathbf{x}_{ref}$ and $\mathbf{x}_{0|t}$ to the same feature space. 

\vspace{3pt}
\noindent \ding{113}~\textbf{Multi Condition Guidance.}
In some more involved applications, multiple conditions can be available to provide control over the generated results. Take the image style transfer task as an example. Here, we have two conditions: the structure information from the source image and the style information from the style image. In these multi-condition cases, assume that the given conditions are denoted as $\{\mathbf{c}_1, \cdots, \mathbf{c}_n\}$, we can approximately construct the energy function as :
\begin{align}
    \mathcal{E}&(\{\mathbf{c}_1, \cdots, \mathbf{c}_n\}, \mathbf{x}_t) \notag \\
    &\approx\eta_1\mathcal{D}_{\boldsymbol{\theta}_1}(\mathbf{c}_1, \mathbf{x}_{0|t})+\cdots+\eta_n\mathcal{D}_{\boldsymbol{\theta}_n}(\mathbf{c}_n, \mathbf{x}_{0|t}),
    \label{eq:multi_condition}
\end{align}
where $\eta_i$ is the weighting factor. We use different distance measuring functions $\{\mathcal{D}_{\boldsymbol{\theta}_1}(\cdot, \cdot), \cdots, \mathcal{D}_{\boldsymbol{\theta}_n}(\cdot, \cdot)\}$ for specific conditions and sum the whole for gradient computation.

\vspace{3pt}
\noindent \ding{113}~\textbf{Guidance for Latent Diffusion.}
Our method applies not only to image diffusions but also to latent diffusions, such as Stable Diffusion~\cite{rombach2022high}. 
In this case, the intermediate results $\mathbf{x}_t$ are latent codes rather than images. We can use the latent decoder to project the generated latent codes to images and then use the same algorithm in the image domain.

\subsection{Specific Examples of Supported Conditions}
\label{subsec:specific_examples}
\vspace{3pt}
\noindent \ding{113}~\textbf{Text.}
For given text prompts, we construct the distance measuring function based on CLIP~\cite{radford2021learning}. Specifically, we take the CLIP image encoder (as $\mathcal{P}_{\boldsymbol{\theta}_2}(\cdot)$) and the CLIP text encoder (as $\mathcal{P}_{\boldsymbol{\theta}_1}(\cdot)$) to project the image $\mathbf{x}_{0|t}$ and given text in the same CLIP feature space. 
Compared with the commonly used cosine distance measurement and for simplicity, we choose the $\ell_2$ Euclidean distance measurement, since the sampling quality in our experiments is not significantly different.

\begin{figure}[!tbp]
  \centering
  \vspace{-0.1cm}
  \includegraphics[width=1\linewidth]{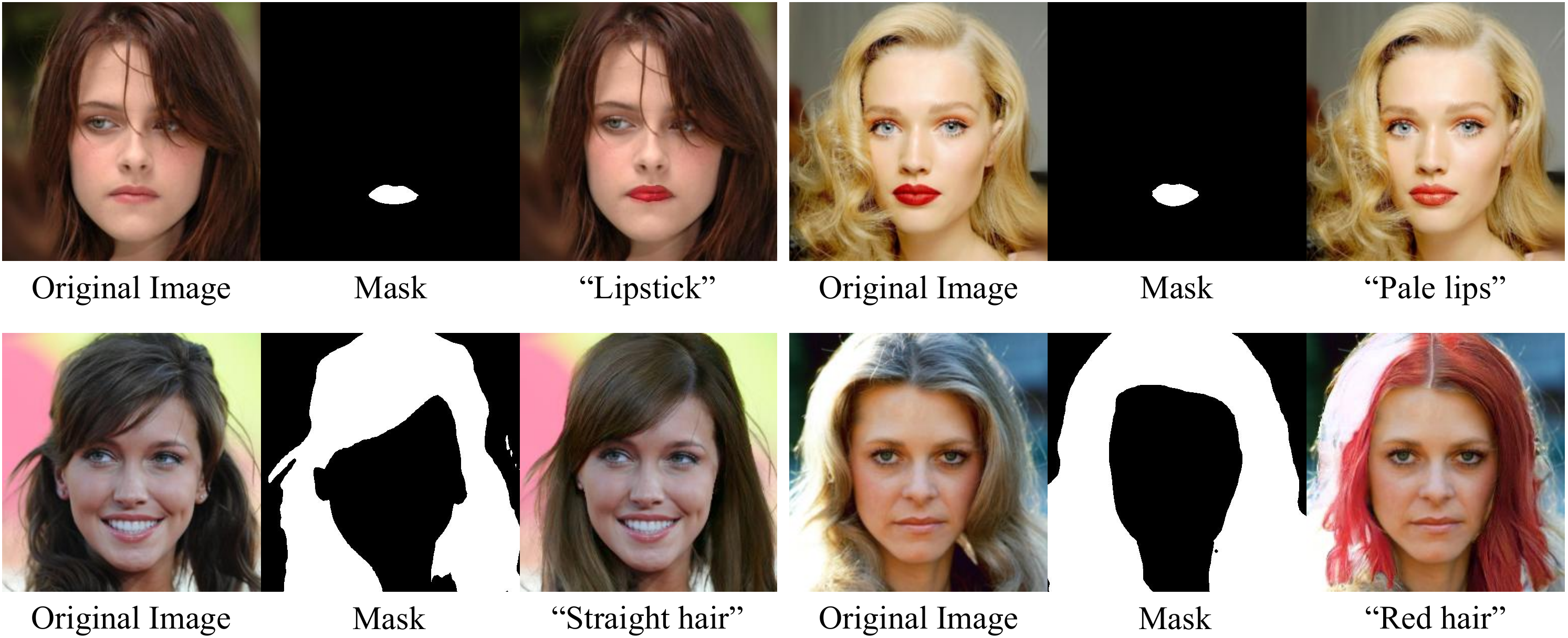}
  \vspace{-0.7cm}
  \caption{\textbf{Practical usage of face parsing maps.} We can limit the gradient of the energy function to update the image only in the target semantic region indicated by the mask so that other regions remain unchanged while editing.}
  \vspace{-0.4cm}
\label{fig:trick_segmentation} 
\end{figure}

\vspace{3pt}
\noindent\ding{113}~\textbf{Segmentation Maps.}
For segmentation maps, we choose a face parsing network based on the real-time semantic segmentation network BiSeNet~\cite{yu2018bisenet} to generate the parsing map of an input human face and directly compute the $\ell_2$ Euclidean distance between the given parsing map and the parsing results of $\mathbf{x}_{0|t}$. An interesting usage of the face parsing network is to constrain the gradient update region so that we can edit the target semantic region without changing other regions (shown in Fig.~\ref{fig:trick_segmentation}).

\vspace{3pt}
\noindent\ding{113}~\textbf{Sketches.}
We choose an open-source pre-trained network~\cite{xiang2022adversarial} that transfers a given anime image to the style of hand-drawn sketches. Experiments prove that the network is still effective for real-world images. We use the $\ell_2$ Euclidean distance to compare the given sketches with transferred sketch-style results of $\mathbf{x}_{0|t}$. 

\vspace{3pt}
\noindent\ding{113}~\textbf{Landmarks.}
We use an open-source pre-trained human face landmark detection network~\cite{PFL} for this application. The detection network has two stages: the first stage finds the position of the center of a face and the second stage marks the landmarks of this detected face. We compute the $\ell_2$ Euclidean distance between predicted face landmarks of $\mathbf{x}_{0|t}$ and the given landmarks condition, and only use the gradient in the face area detected in the first stage to update the intermediate results. 

\vspace{3pt}
\noindent\ding{113}~\textbf{Face IDs.}
We use ArcFace~\cite{deng2019arcface}, an open-source pre-trained human face recognition network, to extract the target features of reference faces to represent face IDs and compute the $\ell_2$ Euclidean distance between the extracted ID features of $\mathbf{x}_{0|t}$ and those of the reference image.

\vspace{3pt}
\noindent\ding{113}~\textbf{Style Images.}
The style image is denoted as $\mathbf{x}_{style}$. We use the following equation to compute the distance of the style information between $\mathbf{x}_{style}$ and $\mathbf{x}_{0|t}$:
\begin{equation}
    Dist(\mathbf{x}_{style}, \mathbf{x}_{0|t}) = ||G(\mathbf{x}_{style})_j - G(\mathbf{x}_{0|t})_j||_F^2,
\end{equation}
where $G(\cdot)_j$ denotes the Gram matrix~\cite{johnson2016perceptual} of the $j$-th layer feature map of an image encoder. In our experiments, we choose the features from the third layer of the CLIP image encoder to generate  satisfactory results.

\vspace{3pt}
\noindent\ding{113}~\textbf{Low-pass Filters.}
For the image transferring task, we need an energy function to constrain the generated results conforming to the structure information of the source image $\mathbf{x}_{source}$. Similar to EGSDE~\cite{zhao2022egsde} and ILVR~\cite{Choi_2021_ICCV}, we choose a low-pass filter $\mathcal{K}(\cdot)$ in this setup. The distance between the source image $\mathbf{x}_{source}$ and $\mathbf{x}_{0|t}$ is computed as:
\begin{equation}
    Dist(\mathbf{x}_{source}, \mathbf{x}_{0|t}) = ||\mathcal{K}(\mathbf{x}_{source}) - \mathcal{K}(\mathbf{x}_{0|t})||^2_2.
\end{equation}


\section{Experiments}

\begin{figure*}[t]
  \centering
  \includegraphics[width=1\linewidth]{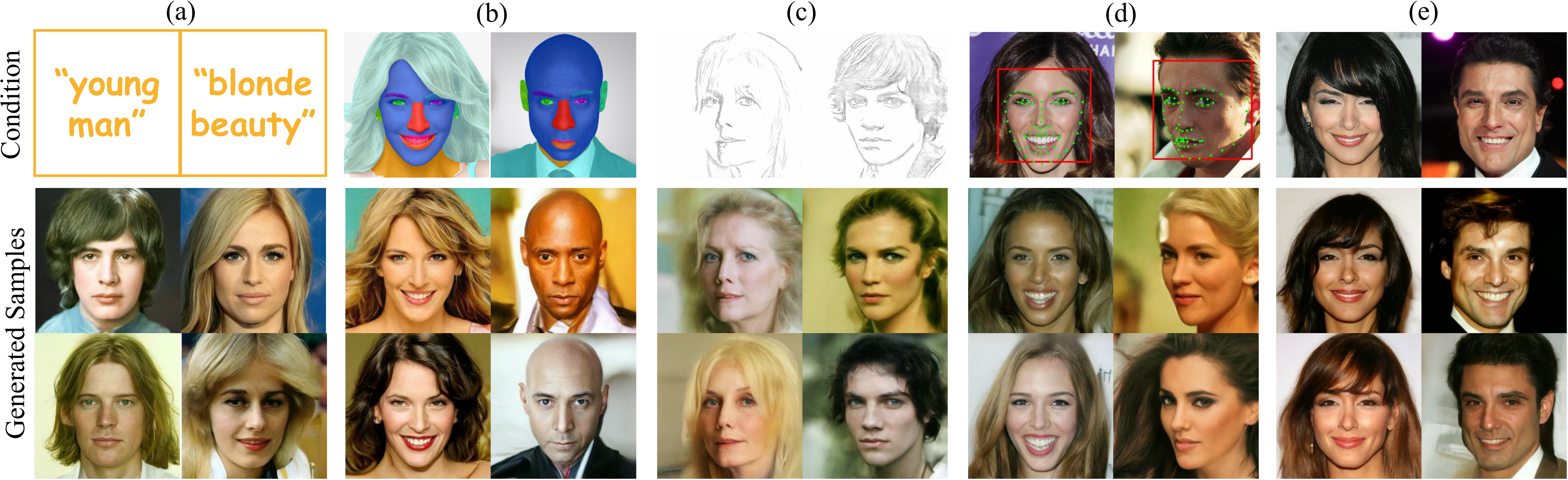}
  \vspace{-0.6cm}
  \caption{\textbf{Qualitative results of using a single condition for human face images.} The included conditions are: (a) text; (b) face parsing maps; (c) sketches; (d) face landmarks; (e) IDs of reference images. \textbf{Zoom in for best view.}}
  \vspace{-0.5cm}
\label{fig:single_energy_face} 
\end{figure*}

\begin{figure}[t]
  \centering
  \vspace{-0.1cm}
  \includegraphics[width=1\linewidth]{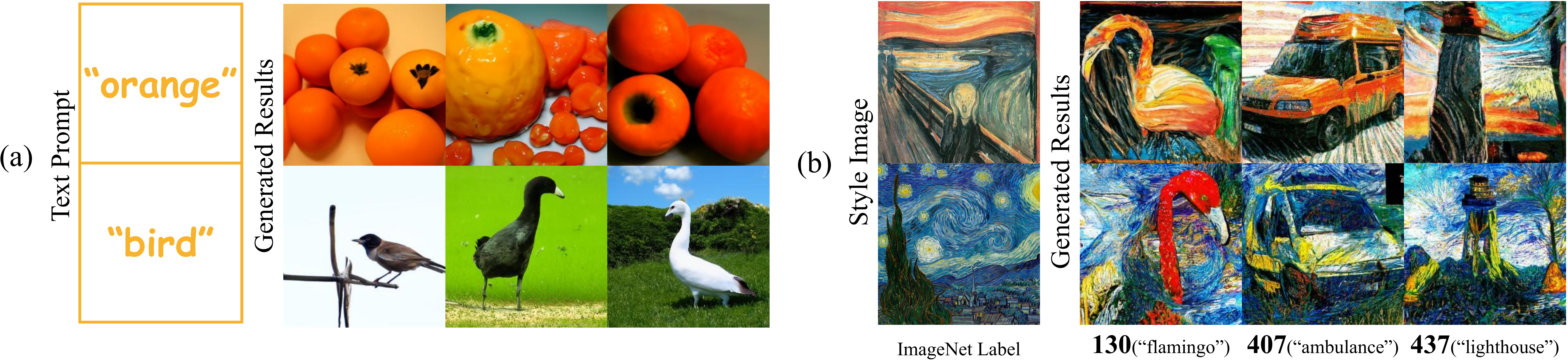}
  \vspace{-0.6cm}
  \caption{\textbf{Qualitative results of using a single condition for ImageNet images.} Pre-trained diffusion models are: (a) unconditional ImageNet diffusion model; (b) classifier-based ImageNet diffusion model. \textbf{Zoom in for best view.}}
  \vspace{-0.3cm}
\label{fig:single_energy_imagenet} 
\end{figure}

\subsection{Implementation Details}
Our proposed method applies to many open-source pre-trained diffusion models (DMs). In our experiment, we have tried the following models and conditions:

\noindent\ding{226}~\textbf{Unconditional Human Face Diffusion Model~\cite{meng2022sdedit}.}
The supported image resolution of this model is $256 \times 256$, and the pre-trained dataset is aligned human faces from CelebA-HQ~\cite{karras2017progressive}. We experiment with conditions that include text, parsing maps, sketches, landmarks, and face IDs. 

\noindent\ding{226}~\textbf{Unconditional ImageNet Diffusion Model~\cite{dhariwal2021diffusion}.}
The supported image resolution of this model is $256 \times 256$ and the pre-trained dataset is ImageNet. We experiment with conditions that include text and style images.  

\noindent\ding{226}~\textbf{Classifier-based ImageNet Diffusion Model~\cite{dhariwal2021diffusion}.}
The supported image resolution of this model is $256 \times 256$, and the pre-trained dataset is ImageNet. This model also has a time-dependent classifier to guide its generation process. We experiment with the condition of style images.

\noindent\ding{226}~\textbf{Stable Diffusion~\cite{rombach2022high}.} Stable Diffusion is a widely used Latent Diffusion Model. The standard resolution of its output images is $512 \times 512$, but it supports  higher resolutions. In our work, we use its pre-trained text-to-image model. We experiment with the condition of style images.

\noindent\ding{226}~\textbf{ControlNet~\cite{zhang2023adding}.} ControlNet is a Stable Diffusion based model supporting extra conditions input with the original text input. In our work, we use the pre-trained pose-to-image and scribble-to-image models. We experiment with conditions that include face IDs and style images.

We choose DDIM~\cite{song2021denoising} with 100 steps as the sampling strategy of all experiments, and other more detailed configurations will be provided in the supplementary material.

\subsection{Qualitative Results}

\noindent\ding{113}~\textbf{Single Condition.}
We present the single-condition-guided results of human face images in Fig.~\ref{fig:single_energy_face}. We can see that the generated results meet the requirements of the given conditions and have rich diversity and good quality. In Fig.~\ref{fig:single_energy_imagenet}, we show the single-condition-guided results of the ImageNet domain. The diversity of the generated results is still high. In order to ensure that the generated results can better meet the control of the given conditions, we use the proposed efficient time-travel strategy.

\begin{figure}[t]
  \centering
  \includegraphics[width=1\linewidth]{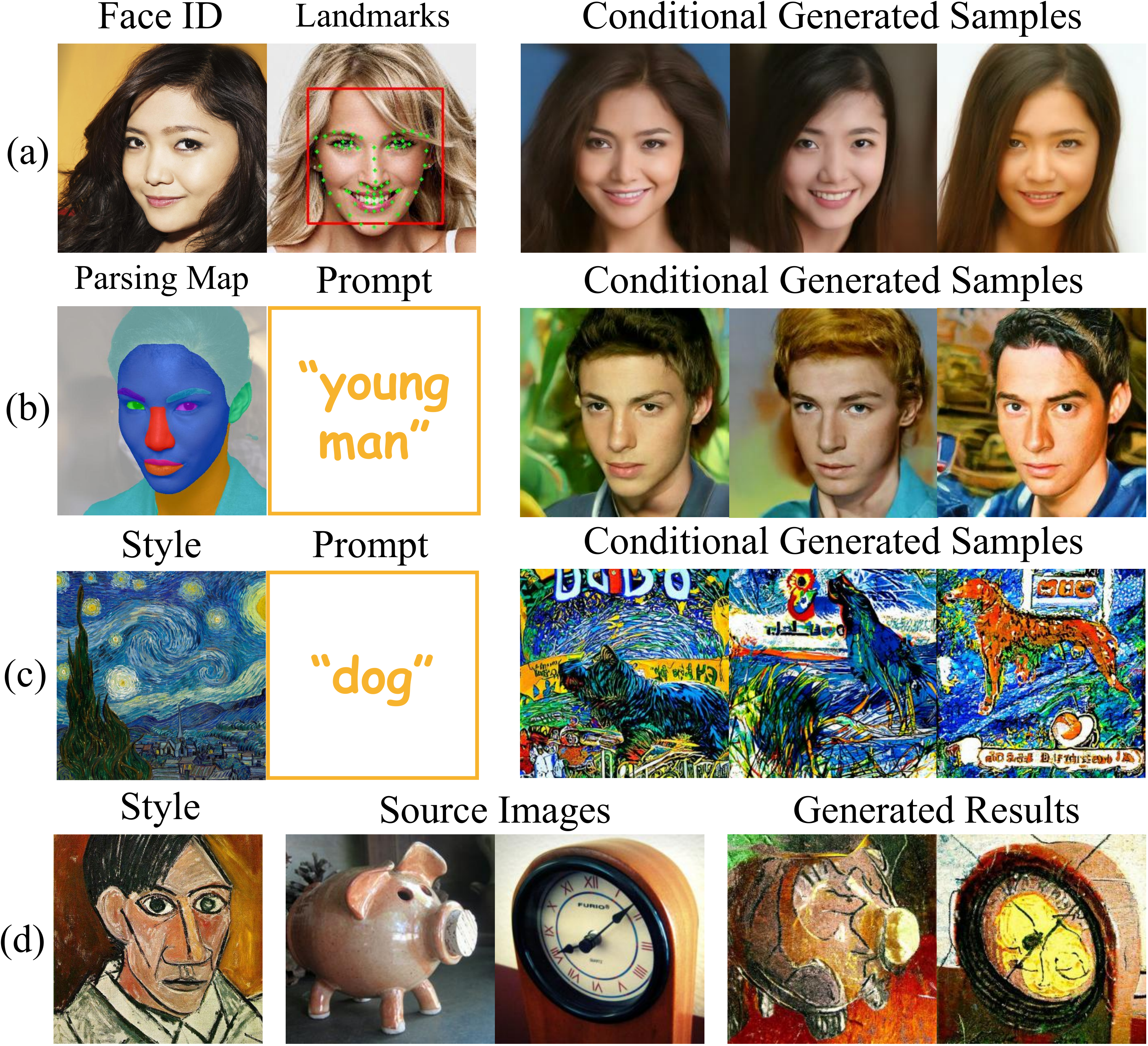}
  \vspace{-0.6cm}
  \caption{\textbf{Qualitative results of using multiple conditions.} Pre-trained models are: (a) and (b): unconditional human face diffusion model; (c) and (d): unconditional ImageNet diffusion model. \textbf{Zoom in for best view.}}
  \vspace{-0.5cm}
\label{fig:multi_energy} 
\end{figure}

\noindent\ding{113}~\textbf{Multiple Conditions.}
Fig.~\ref{fig:multi_energy} shows the synthesized results guided by multiple conditions in the domain of human faces and ImageNet. In the human face domain (a small data domain), we produce good results with rich diversity and high consistency with the conditions. We use the efficient time-travel strategy in the ImageNet domain (a large data domain) to produce acceptable results.

\noindent\ding{113}~\textbf{Training-free Guidance for Latent Domain.} It should be pointed out that FreeDoM supports diffusion models in both image and latent domains. In our work, we experiment with two latent diffusion models: Stable Diffusion~\cite{rombach2022high} and ControlNet~\cite{zhang2023adding}. 
We try to add the training-free conditional interfaces based on their energy functions to work with the existing training-required conditional interfaces, leading to satisfactory results  shown in Fig.~\ref{fig:overview}(d)-(f). As such, we can see great application potential for mixing training-free and training-required conditional interfaces in various practical applications.

\subsection{Further Studies}

\begin{figure*}[t]
  \centering
  \includegraphics[width=1\linewidth]{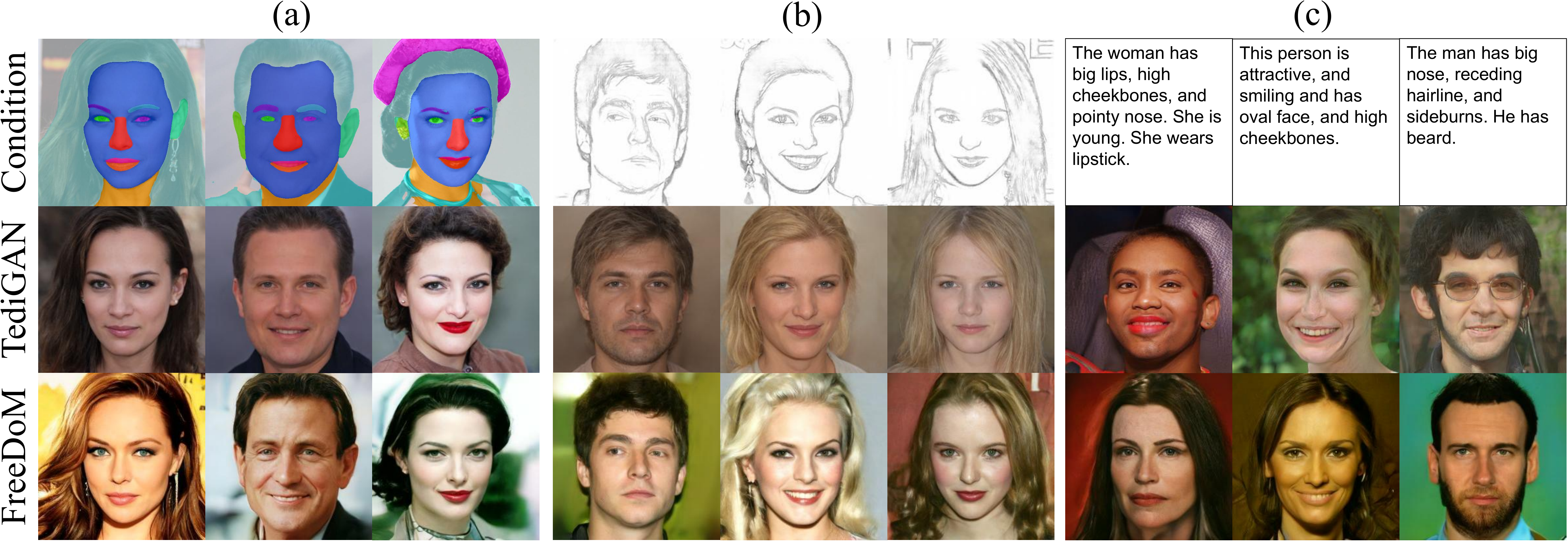}
  \vspace{-0.6cm}
  \caption{\textbf{Comparison between FreeDoM and TediGAN~\cite{xia2021tedigan} in three conditional image synthesis tasks}: (a) segmentation maps to human faces; (b) sketches to human faces; (c) text prompts to human faces. \textbf{Zoom in for best view.}}
  \vspace{-0.5cm}
\label{fig:comp_tedigan} 
\end{figure*}

\begin{table}[t]
    \centering
    \scriptsize
    \tabcolsep=0.09cm
    \begin{tabular}{c | cc | cc | cc}
        \toprule[1.2pt]
          \multirow{2}{*}{Methods} & \multicolumn{2}{c}{Segmentation maps} \vline & \multicolumn{2}{c}{Sketches}\vline & \multicolumn{2}{c}{Texts}\\
        \rule{0pt}{8pt}
            & Distance↓ & FID↓ & Distance↓ & FID↓ & Distance↓ & FID↓\\
        \toprule[1.2pt]
TediGAN~\cite{xia2021tedigan} & 2037.2 & \textbf{52.77} & 48.61 & 91.11 & 12.31 & 71.71\\
\rule{0pt}{10pt}
        FreeDoM (ours) & \textbf{1696.1} & 53.08 & \textbf{33.29} & \textbf{70.97} & \textbf{10.83} & \textbf{55.91} \\
        \bottomrule[1.2pt]
    \end{tabular}
    \caption{We compare FreeDoM with the training-required method TediGAN~\cite{xia2021tedigan} in three image conditional synthesis tasks. We compute the distance with given conditions and FID to judge the performance. The comparison shows that FreeDoM generates images matching given conditions better and having a comparable or better image quality.
    }
    \vspace{-0.5cm}
    \label{tb:comp_tedigan}
\end{table}

\noindent\ding{113}~\textbf{Comparison between FreeDoM and TediGAN~\cite{xia2021tedigan}.}
We compare FreeDoM with the training-required conditional human face generation method TediGAN under three conditions: segmentation maps, sketches, and text. 
A qualitative comparison is shown in Fig.~\ref{fig:comp_tedigan}, and quantitative comparison results are reported in Tab.~\ref{tb:comp_tedigan}.
For the comparison, we choose 1000 segmentation maps, 1000 sketches, and 1000 text prompts to generate 1000 results, respectively. Then we compute FID and the average distance with given conditions using the methods introduced in Sec.~\ref{subsec:specific_examples} to judge the performance. The comparison shows that the images generated by FreeDoM match the given conditions better and have a comparable or better image quality.

\begin{figure}[t]
  \centering
  \includegraphics[width=1\linewidth]{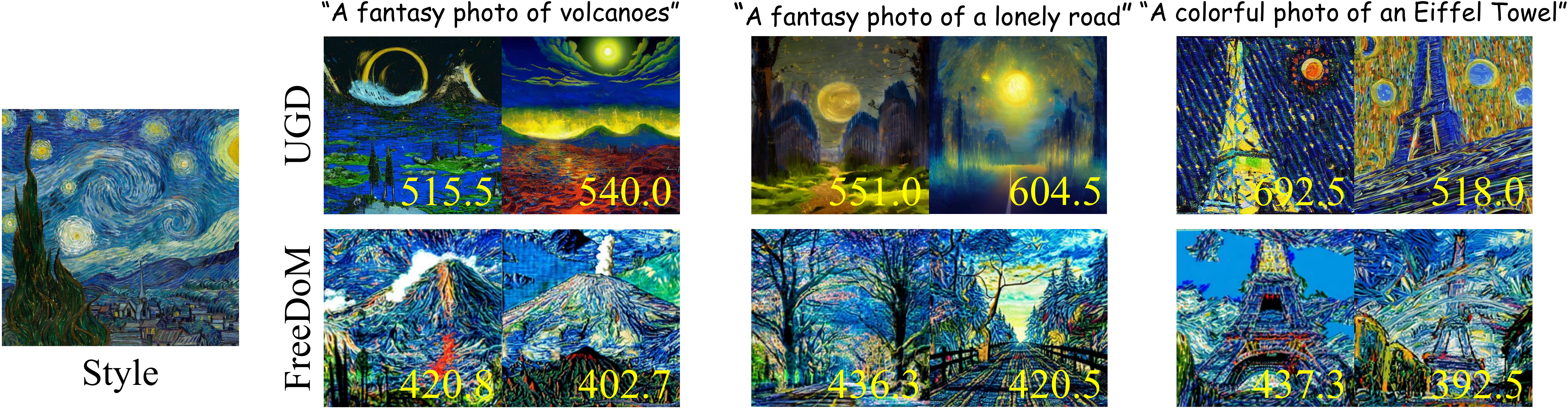}
  \vspace{-0.6cm}
  \caption{\textbf{Comparison between FreeDoM and UGD~\cite{bansal2023universal} in style-guided generation.} 
  The UGD results are taken from the original paper. 
  The number in the lower right corner of each image represents its distance with the provided style image (smaller is better), which is calculated using the method described in Sec.~\ref{subsec:specific_examples}.
  FreeDoM offers obvious advantages in image quality and in the degree of statisfaction of the conditions.
  \textbf{Zoom in for best view.}
  }
  \vspace{-0.3cm}
\label{fig:comp_ugd} 
\end{figure}

\noindent\ding{113}~\textbf{Comparison between FreeDoM and UGD~\cite{bansal2023universal}.}
We compare FreeDoM with Universal Guidance Diffusion (UGD)~\cite{bansal2023universal} in style-guided generations. From Fig.~\ref{fig:comp_ugd}, we find that FreeDoM has significant advantages over UGD in the degree of alignment with the conditioned style image. Regarding the inference speed, UGD runs in about 40 minutes (using the open-source code) on a GeForce RTX 3090 GPU to synthesize one image with a resolution of $512 \times 512$, while we only need about 84 seconds (nearly 30$\times$ faster). 

\begin{figure}[t]
  \centering
  \includegraphics[width=1\linewidth]{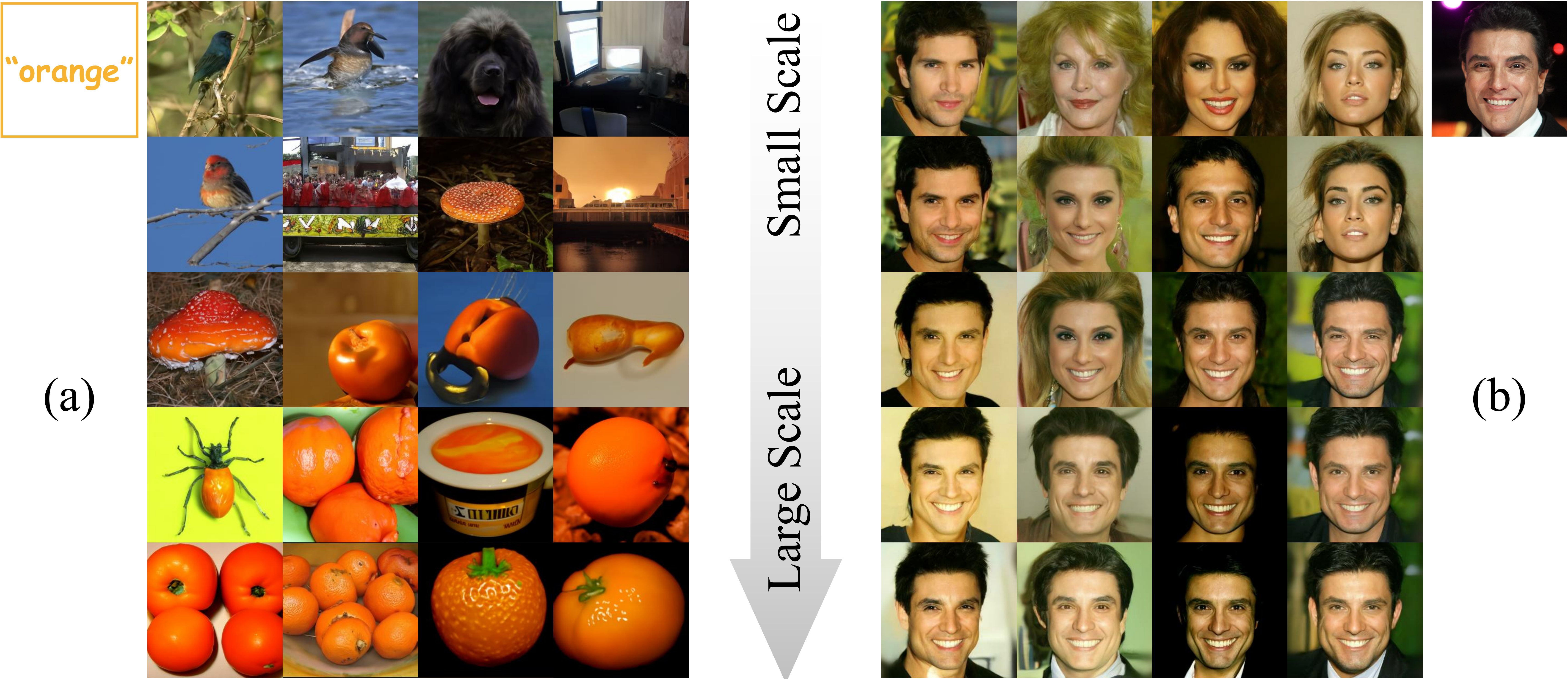}
  \vspace{-0.6cm}
  \caption{\textbf{Demonstration of the effect of different learning rates from small scale to large scale.} (a): unconditional ImageNet diffusion models with prompt ``orange''; (b): unconditional human face diffusion models with a face ID from the reference image. \textbf{Zoom in for best view.}}
  \vspace{-0.5cm}
\label{fig:scalable} 
\end{figure}

\noindent\ding{113}~\textbf{Effect of different learning rates.}
We studied the effect of different learning rates on the results.
Fig.~\ref{fig:scalable} shows the results while increasing the energy function's learning rate ($\rho_t$ in Eq.~(\ref{eq:approximated_energy_guided_sampling})) from $0$. We can see that FreeDoM is scalable in terms of its control ability, which means that users can adjust the intensity of control as needed. 


\section{Conclusions \& Limitations}
We propose a training-free energy-guided conditional diffusion model, FreeDoM, to address a wide range of conditional generation tasks without training. Our method uses off-the-shelf pre-trained time-independent networks to approximate the time-dependent energy functions. Then, we use the gradient of the approximated energy to guide the generation process. Our method supports different diffusion models, including image  and latent diffusion models. It is worth emphasizing that the applications presented in this paper are only a subset of the applications FreeDoM supports and should not be limited to these. In future work, we aim to explore even more energy functions for a broader range of tasks.

Despite its merits, our FreeDoM method has some limitations: (1) The time cost of the sampling is still higher than the training-required methods because each iteration adds a derivative operation for the energy function, and the time-travel strategy introduces more sampling steps. (2)  It is difficult to use the energy function to control the fine-grained structure features in the large data domain.
For example, using the Canny edge maps as the conditions may result in poor guidance, even if we use the time-travel strategy. In this case, the training-required methods will provide a better alternative. (3) Eq.~\ref{eq:multi_condition} deals with multi-condition control and assumes that the provided conditions  are independent, which is not necessarily true in practice. When conditions conflict with each other, FreeDoM may produce subpar generation results.

{\small
\bibliographystyle{ieee_fullname}
\bibliography{egbib}

\begin{thebibliography}{10}\itemsep=-1pt

\bibitem{avrahami2022blended}
Omri Avrahami, Dani Lischinski, and Ohad Fried.
\newblock Blended diffusion for text-driven editing of natural images.
\newblock In {\em Proceedings of the IEEE/CVF Conference on Computer Vision and
  Pattern Recognition}, 2022.

\bibitem{bansal2023universal}
Arpit Bansal, Hong-Min Chu, Avi Schwarzschild, Soumyadip Sengupta, Micah
  Goldblum, Jonas Geiping, and Tom Goldstein.
\newblock Universal guidance for diffusion models.
\newblock {\em arXiv preprint arXiv:2302.07121}, 2023.

\bibitem{brocklarge}
Andrew Brock, Jeff Donahue, and Karen Simonyan.
\newblock Large scale gan training for high fidelity natural image synthesis.
\newblock In {\em International Conference on Learning Representations (ICLR)},
  2019.

\bibitem{PFL}
Cunjian Chen.
\newblock {PyTorch Face Landmark}: A fast and accurate facial landmark
  detector, 2021.

\bibitem{Choi_2021_ICCV}
Jooyoung Choi, Sungwon Kim, Yonghyun Jeong, Youngjune Gwon, and Sungroh Yoon.
\newblock Ilvr: Conditioning method for denoising diffusion probabilistic
  models.
\newblock In {\em Proceedings of the IEEE/CVF International Conference on
  Computer Vision (ICCV)}, 2021.

\bibitem{chung2023diffusion}
Hyungjin Chung, Jeongsol Kim, Michael~Thompson Mccann, Marc~Louis Klasky, and
  Jong~Chul Ye.
\newblock Diffusion posterior sampling for general noisy inverse problems.
\newblock In {\em International Conference on Learning Representations}, 2023.

\bibitem{Chung_2022_CVPR}
Hyungjin Chung, Byeongsu Sim, and Jong~Chul Ye.
\newblock Come-closer-diffuse-faster: Accelerating conditional diffusion models
  for inverse problems through stochastic contraction.
\newblock In {\em Proceedings of the IEEE/CVF Conference on Computer Vision and
  Pattern Recognition (CVPR)}, 2022.

\bibitem{chung2022improving}
Hyungjin Chung, Byeongsu Sim, and Jong~Chul Ye.
\newblock Improving diffusion models for inverse problems using manifold
  constraints.
\newblock In {\em Advances in Neural Information Processing Systems (NeurIPS)},
  2022.

\bibitem{deng2019arcface}
Jiankang Deng, Jia Guo, Niannan Xue, and Stefanos Zafeiriou.
\newblock Arcface: Additive angular margin loss for deep face recognition.
\newblock In {\em Proceedings of the IEEE/CVF conference on computer vision and
  pattern recognition}, 2019.

\bibitem{dhariwal2021diffusion}
Prafulla Dhariwal and Alexander Nichol.
\newblock Diffusion models beat gans on image synthesis.
\newblock In {\em Advances in Neural Information Processing Systems (NeurIPS)},
  2021.

\bibitem{feng2023trainingfree}
Weixi Feng, Xuehai He, Tsu-Jui Fu, Varun Jampani, Arjun~Reddy Akula, Pradyumna
  Narayana, Sugato Basu, Xin~Eric Wang, and William~Yang Wang.
\newblock Training-free structured diffusion guidance for compositional
  text-to-image synthesis.
\newblock In {\em International Conference on Learning Representations}, 2023.

\bibitem{gan}
Ian Goodfellow, Jean Pouget-Abadie, Mehdi Mirza, Bing Xu, David Warde-Farley,
  Sherjil Ozair, Aaron Courville, and Yoshua Bengio.
\newblock Generative adversarial nets.
\newblock {\em Advances in Neural Information Processing Systems (NeurIPS)},
  2014.

\bibitem{gu2021vector}
Shuyang Gu, Dong Chen, Jianmin Bao, Fang Wen, Bo Zhang, Dongdong Chen, Lu Yuan,
  and Baining Guo.
\newblock Vector quantized diffusion model for text-to-image synthesis.
\newblock In {\em Proceedings of the IEEE/CVF Conference on Computer Vision and
  Pattern Recognition (CVPR)}, 2022.

\bibitem{hertz2022prompt}
Amir Hertz, Ron Mokady, Jay Tenenbaum, Kfir Aberman, Yael Pritch, and Daniel
  Cohen-Or.
\newblock Prompt-to-prompt image editing with cross attention control.
\newblock {\em arXiv preprint arXiv:2208.01626}, 2022.

\bibitem{ho2020denoising}
Jonathan Ho, Ajay Jain, and Pieter Abbeel.
\newblock Denoising diffusion probabilistic models.
\newblock In {\em Advances in Neural Information Processing Systems (NeurIPS)},
  2020.

\bibitem{ho2022classifier}
Jonathan Ho and Tim Salimans.
\newblock Classifier-free diffusion guidance.
\newblock In {\em NeurIPS 2021 Workshop on Deep Generative Models and
  Downstream Applications}.

\bibitem{johnson2016perceptual}
Justin Johnson, Alexandre Alahi, and Li Fei-Fei.
\newblock Perceptual losses for real-time style transfer and super-resolution.
\newblock In {\em Proceedings of the European conference on computer vision
  (ECCV)}, 2016.

\bibitem{karras2017progressive}
Tero Karras, Timo Aila, Samuli Laine, and Jaakko Lehtinen.
\newblock Progressive growing of gans for improved quality, stability, and
  variation.
\newblock {\em Internation Conference on Reoresentation Learning (ICLR)}, 2018.

\bibitem{kawar2022denoising}
Bahjat Kawar, Michael Elad, Stefano Ermon, and Jiaming Song.
\newblock Denoising diffusion restoration models.
\newblock In {\em Advances in Neural Information Processing Systems (NeurIPS)},
  2022.

\bibitem{Kim_2022_CVPR}
Gwanghyun Kim, Taesung Kwon, and Jong~Chul Ye.
\newblock Diffusionclip: Text-guided diffusion models for robust image
  manipulation.
\newblock In {\em Proceedings of the IEEE/CVF Conference on Computer Vision and
  Pattern Recognition (CVPR)}, 2022.

\bibitem{lecun2006tutorial}
Yann LeCun, Sumit Chopra, Raia Hadsell, M Ranzato, and Fujie Huang.
\newblock A tutorial on energy-based learning.
\newblock {\em Predicting structured data}, 1(0), 2006.

\bibitem{liu2022compositional}
Nan Liu, Shuang Li, Yilun Du, Antonio Torralba, and Joshua~B Tenenbaum.
\newblock Compositional visual generation with composable diffusion models.
\newblock In {\em Proceedings of the European Conference on Computer Vision
  (ECCV)}, 2022.

\bibitem{liu2019more}
Xihui Liu, Dong~Huk Park, Samaneh Azadi, Gong Zhang, Arman Chopikyan, Yuxiao
  Hu, Humphrey Shi, Anna Rohrbach, and Trevor Darrell.
\newblock More control for free! image synthesis with semantic diffusion
  guidance.
\newblock In {\em Proceedings of the IEEE/CVF Winter Conference on Applications
  of Computer Vision}, 2023.

\bibitem{Lugmayr_2022_CVPR}
Andreas Lugmayr, Martin Danelljan, Andres Romero, Fisher Yu, Radu Timofte, and
  Luc Van~Gool.
\newblock Repaint: Inpainting using denoising diffusion probabilistic models.
\newblock In {\em Proceedings of the IEEE/CVF Conference on Computer Vision and
  Pattern Recognition (CVPR)}, 2022.

\bibitem{meng2022sdedit}
Chenlin Meng, Yutong He, Yang Song, Jiaming Song, Jiajun Wu, Jun-Yan Zhu, and
  Stefano Ermon.
\newblock {SDE}dit: Guided image synthesis and editing with stochastic
  differential equations.
\newblock In {\em International Conference on Learning Representations}, 2022.

\bibitem{mirza2014conditional}
Mehdi Mirza and Simon Osindero.
\newblock Conditional generative adversarial nets.
\newblock {\em arXiv preprint arXiv:1411.1784}, 2014.

\bibitem{ron2022nulltext}
Ron Mokady, Amir Hertz, Kfir Aberman, Yael Pritch, and Daniel Cohen-Or.
\newblock Null-text inversion for editing real images using guided diffusion
  models.
\newblock {\em arXiv preprint arXiv:2211.09794}, 2022.

\bibitem{mou2023adapter}
Chong Mou, Xintao Wang, Liangbin Xie, Jian Zhang, Zhongang Qi, Ying Shan, and
  Xiaohu Qie.
\newblock T2i-adapter: Learning adapters to dig out more controllable ability
  for text-to-image diffusion models.
\newblock {\em arXiv preprint arXiv:2302.08453}, 2023.

\bibitem{nichol2021glide}
Alex Nichol, Prafulla Dhariwal, Aditya Ramesh, Pranav Shyam, Pamela Mishkin,
  Bob McGrew, Ilya Sutskever, and Mark Chen.
\newblock Glide: Towards photorealistic image generation and editing with
  text-guided diffusion models.
\newblock {\em arXiv preprint arXiv:2112.10741}, 2021.

\bibitem{parmar2023zero}
Gaurav Parmar, Krishna~Kumar Singh, Richard Zhang, Yijun Li, Jingwan Lu, and
  Jun-Yan Zhu.
\newblock Zero-shot image-to-image translation.
\newblock {\em arXiv preprint arXiv:2302.03027}, 2023.

\bibitem{radford2021learning}
Alec Radford, Jong~Wook Kim, Chris Hallacy, Aditya Ramesh, Gabriel Goh,
  Sandhini Agarwal, Girish Sastry, Amanda Askell, Pamela Mishkin, Jack Clark,
  et~al.
\newblock Learning transferable visual models from natural language
  supervision.
\newblock In {\em International Conference on Machine Learning (ICML)}. PMLR,
  2021.

\bibitem{ramesh2022hierarchical}
Aditya Ramesh, Prafulla Dhariwal, Alex Nichol, Casey Chu, and Mark Chen.
\newblock Hierarchical text-conditional image generation with clip latents.
\newblock {\em arXiv preprint arXiv:2204.06125}, 2022.

\bibitem{rombach2022high}
Robin Rombach, Andreas Blattmann, Dominik Lorenz, Patrick Esser, and Bj{\"o}rn
  Ommer.
\newblock High-resolution image synthesis with latent diffusion models.
\newblock In {\em Proceedings of the IEEE/CVF Conference on Computer Vision and
  Pattern Recognition (CVPR)}, 2022.

\bibitem{saharia2022palette}
Chitwan Saharia, William Chan, Huiwen Chang, Chris Lee, Jonathan Ho, Tim
  Salimans, David Fleet, and Mohammad Norouzi.
\newblock Palette: Image-to-image diffusion models.
\newblock In {\em ACM SIGGRAPH 2022 Conference Proceedings}, 2022.

\bibitem{saharia2022photorealistic}
Chitwan Saharia, William Chan, Saurabh Saxena, Lala Li, Jay Whang, Emily
  Denton, Seyed Kamyar~Seyed Ghasemipour, Raphael Gontijo-Lopes, Burcu~Karagol
  Ayan, Tim Salimans, Jonathan Ho, David~J. Fleet, and Mohammad Norouzi.
\newblock Photorealistic text-to-image diffusion models with deep language
  understanding.
\newblock In Alice~H. Oh, Alekh Agarwal, Danielle Belgrave, and Kyunghyun Cho,
  editors, {\em Advances in Neural Information Processing Systems}, 2022.

\bibitem{saharia2022image}
Chitwan Saharia, Jonathan Ho, William Chan, Tim Salimans, David~J Fleet, and
  Mohammad Norouzi.
\newblock Image super-resolution via iterative refinement.
\newblock {\em IEEE Transactions on Pattern Analysis and Machine Intelligence},
  2022.

\bibitem{sheynin2023knndiffusion}
Shelly Sheynin, Oron Ashual, Adam Polyak, Uriel Singer, Oran Gafni, Eliya
  Nachmani, and Yaniv Taigman.
\newblock {KNN}-diffusion: Image generation via large-scale retrieval.
\newblock In {\em International Conference on Learning Representations}, 2023.

\bibitem{song2021denoising}
Jiaming Song, Chenlin Meng, and Stefano Ermon.
\newblock Denoising diffusion implicit models.
\newblock In {\em International Conference on Learning Representations (ICLR)},
  2021.

\bibitem{song2023pseudoinverseguided}
Jiaming Song, Arash Vahdat, Morteza Mardani, and Jan Kautz.
\newblock Pseudoinverse-guided diffusion models for inverse problems.
\newblock In {\em International Conference on Learning Representations}, 2023.

\bibitem{song2019generative}
Yang Song and Stefano Ermon.
\newblock Generative modeling by estimating gradients of the data distribution.
\newblock In {\em Advances in Neural Information Processing Systems (NeurIPS)},
  2019.

\bibitem{song2021solving}
Yang Song, Liyue Shen, Lei Xing, and Stefano Ermon.
\newblock Solving inverse problems in medical imaging with score-based
  generative models.
\newblock In {\em International Conference on Learning Representations (ICLR)},
  2021.

\bibitem{song2021scorebased}
Yang Song, Jascha Sohl-Dickstein, Diederik~P Kingma, Abhishek Kumar, Stefano
  Ermon, and Ben Poole.
\newblock Score-based generative modeling through stochastic differential
  equations.
\newblock In {\em International Conference on Learning Representations (ICLR)},
  2021.

\bibitem{wang2023unlimited}
Yinhuai Wang, Jiwen Yu, Runyi Yu, and Jian Zhang.
\newblock Unlimited-size diffusion restoration.
\newblock {\em arXiv preprint arXiv:2303.00354}, 2023.

\bibitem{wang2023zeroshot}
Yinhuai Wang, Jiwen Yu, and Jian Zhang.
\newblock Zero-shot image restoration using denoising diffusion null-space
  model.
\newblock In {\em International Conference on Learning Representations}, 2023.

\bibitem{whang2022deblurring}
Jay Whang, Mauricio Delbracio, Hossein Talebi, Chitwan Saharia, Alexandros~G
  Dimakis, and Peyman Milanfar.
\newblock Deblurring via stochastic refinement.
\newblock In {\em Proceedings of the IEEE/CVF Conference on Computer Vision and
  Pattern Recognition (CVPR)}, 2022.

\bibitem{xia2021tedigan}
Weihao Xia, Yujiu Yang, Jing-Hao Xue, and Baoyuan Wu.
\newblock Tedigan: Text-guided diverse face image generation and manipulation.
\newblock In {\em IEEE Conference on Computer Vision and Pattern Recognition
  (CVPR)}, 2021.

\bibitem{xiang2022adversarial}
Xiaoyu Xiang, Ding Liu, Xiao Yang, Yiheng Zhu, Xiaohui Shen, and Jan~P
  Allebach.
\newblock Adversarial open domain adaptation for sketch-to-photo synthesis.
\newblock In {\em Proceedings of the IEEE/CVF Winter Conference on Applications
  of Computer Vision}, 2022.

\bibitem{yu2018bisenet}
Changqian Yu, Jingbo Wang, Chao Peng, Changxin Gao, Gang Yu, and Nong Sang.
\newblock Bisenet: Bilateral segmentation network for real-time semantic
  segmentation.
\newblock In {\em Proceedings of the European conference on computer vision
  (ECCV)}, 2018.

\bibitem{zhang2023adding}
Lvmin Zhang and Maneesh Agrawala.
\newblock Adding conditional control to text-to-image diffusion models.
\newblock {\em arXiv preprint arXiv:2302.05543}, 2023.

\bibitem{zhao2022egsde}
Min Zhao, Fan Bao, Chongxuan Li, and Jun Zhu.
\newblock Egsde: Unpaired image-to-image translation via energy-guided
  stochastic differential equations.
\newblock {\em Advances in Neural Information Processing Systems (NeurIPS)},
  2022.

\end{thebibliography}
}

\appendix
\onecolumn{
\begin{appendices}

This appendix is organized as follows:
\begin{itemize}
    \item Section~\ref{sec:more_results}: More results to show the performance of FreeDoM.
    \item Section~\ref{sec:zir}: The relationship between FreeDoM and zero-shot image restoration methods.
\end{itemize}

\section{More Results}
\label{sec:more_results}
In this section, we provide more generated results to demonstrate the effects of FreeDoM under various conditions and the applications FreeDoM support with training-required latent diffusion models. 

We show the results of various conditions in Fig.~\ref{fig:text2face} (text-to-image), Fig.~\ref{fig:seg2face} (segmentation-to-image), Fig.~\ref{fig:sketch2face} (sketch-to-image), Fig.~\ref{fig:land2face} (landmark-to-image), and Fig.~\ref{fig:id2face} (id-to-image).

We show the results with latent diffusion models in Fig.~\ref{fig:sd_style} (style guidance $+$ Stable Diffusion~\cite{rombach2022high}), Fig.~\ref{fig:cn_style} (style guidance $+$ Scribble ControlNet~\cite{zhang2023adding}) and Fig.~\ref{fig:cn_id} (face ID guidance $+$ Human-pose ControlNet~\cite{zhang2023adding}).
In order to further illustrate the implementation process of the application with the Human-pose ControlNet demonstrated in Fig.~\ref{fig:cn_id}, we provide Fig.~\ref{fig:cn_id_process}.

\begin{figure*}[ht]
  \centering
\includegraphics[width=1\linewidth]{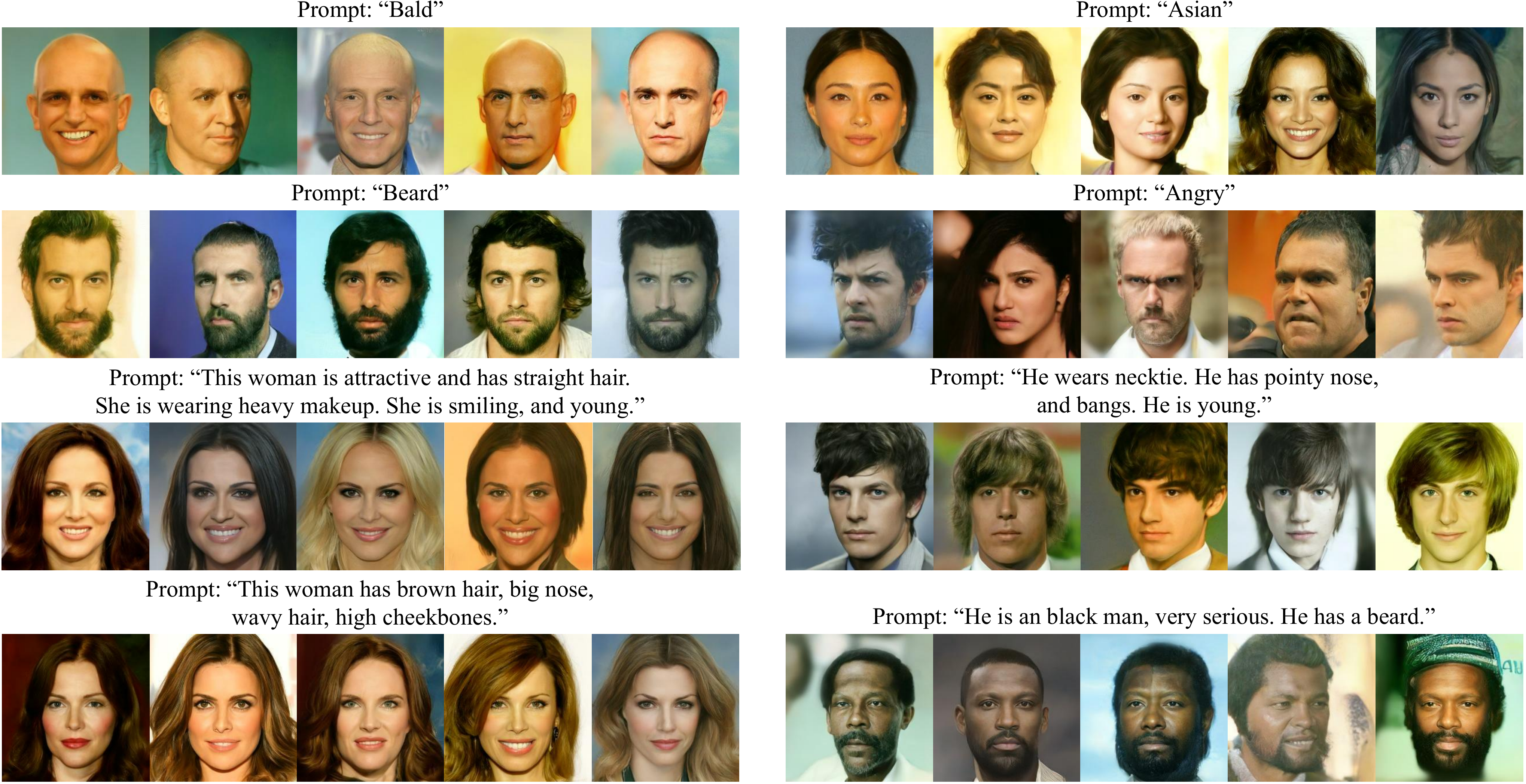}
  \caption{Generated human faces for the text-to-image task. We choose four short and four long prompts to demonstrate the performance of FreeDoM. The characteristics described by these short prompts are experientially seldom seen in the training set. These results are consistent with the given conditions and have good diversity.}
\label{fig:text2face} 
\end{figure*}
\clearpage

\begin{figure*}[ht]
  \centering
\includegraphics[width=1\linewidth]{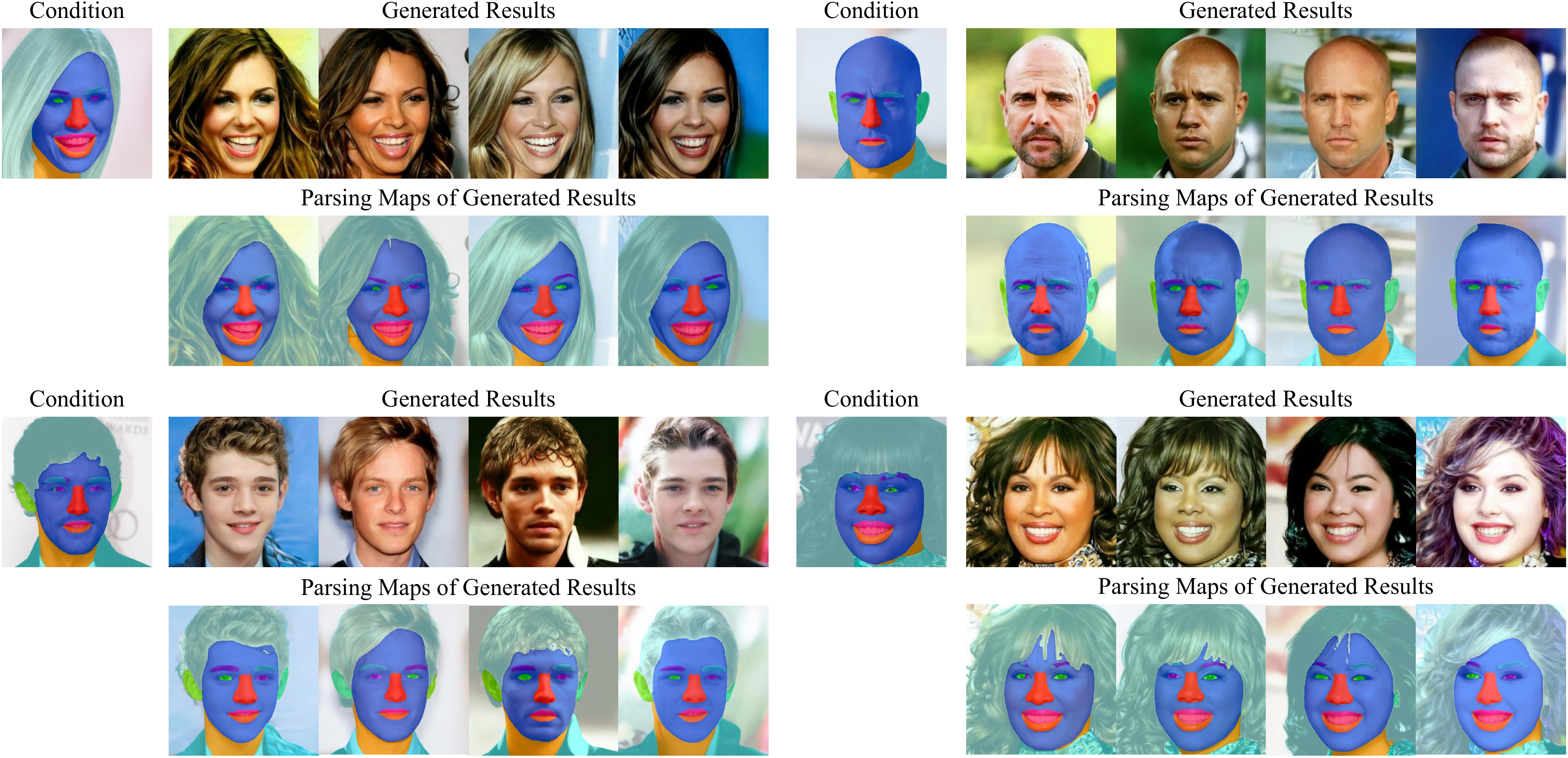}
  \caption{Generated human faces for the segmentation-to-image task. We choose four parsing maps to guide the generation process and output the parsing maps of the generated results to check the matching degree with given conditions. We can see that these results are consistent with the given conditions and have good diversity.}
\label{fig:seg2face} 
\end{figure*}

\begin{figure*}[ht]
  \centering
\includegraphics[width=1\linewidth]{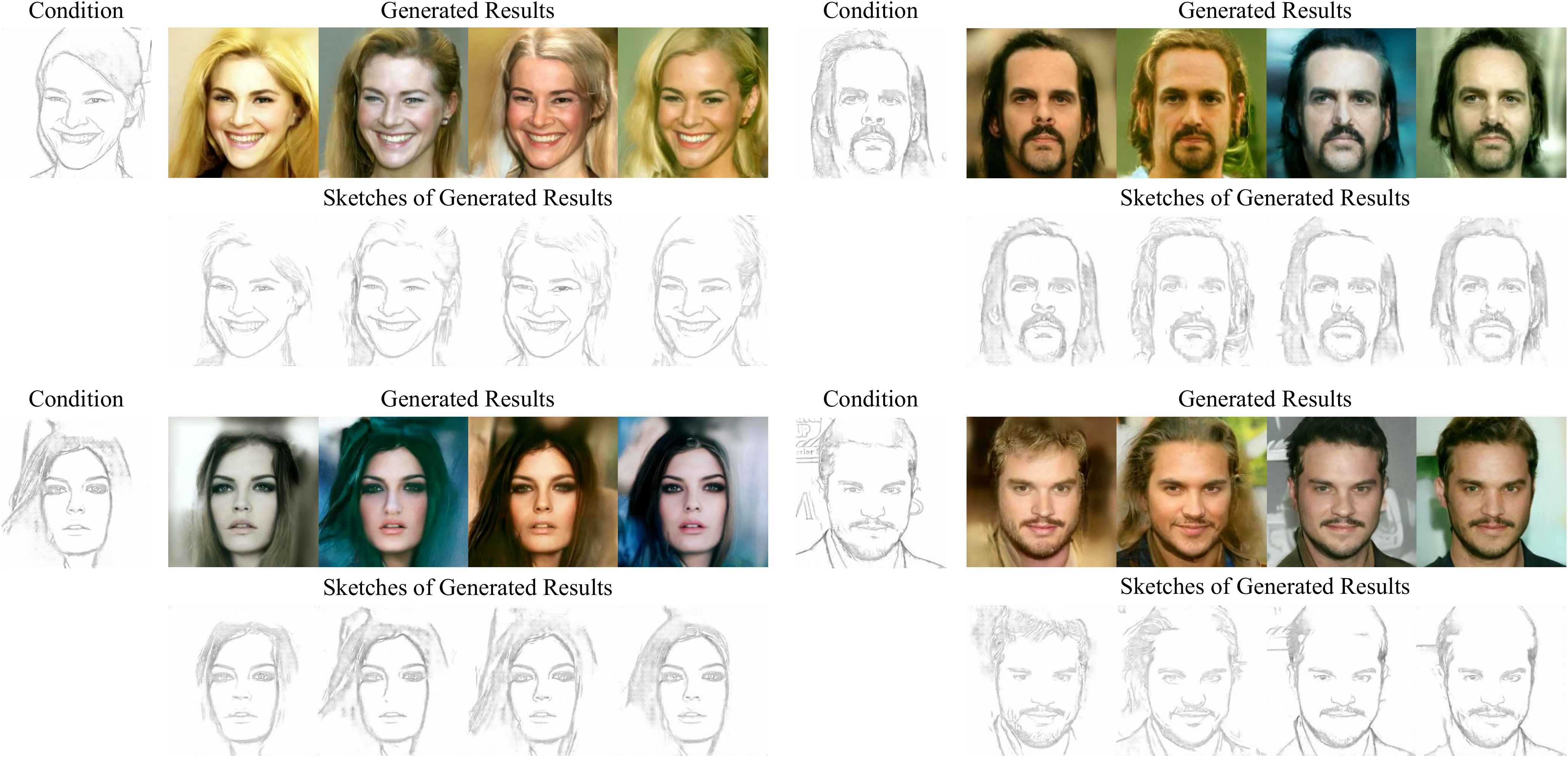}
  \caption{Generated human faces for the sketch-to-image task. We choose four sketches to guide the generation process and output the sketches of the generated results to check the matching degree with the given conditions. These results are consistent with the given conditions and have good diversity.}
\label{fig:sketch2face} 
\end{figure*}
\clearpage

\begin{figure*}[ht]
  \centering
\includegraphics[width=1\linewidth]{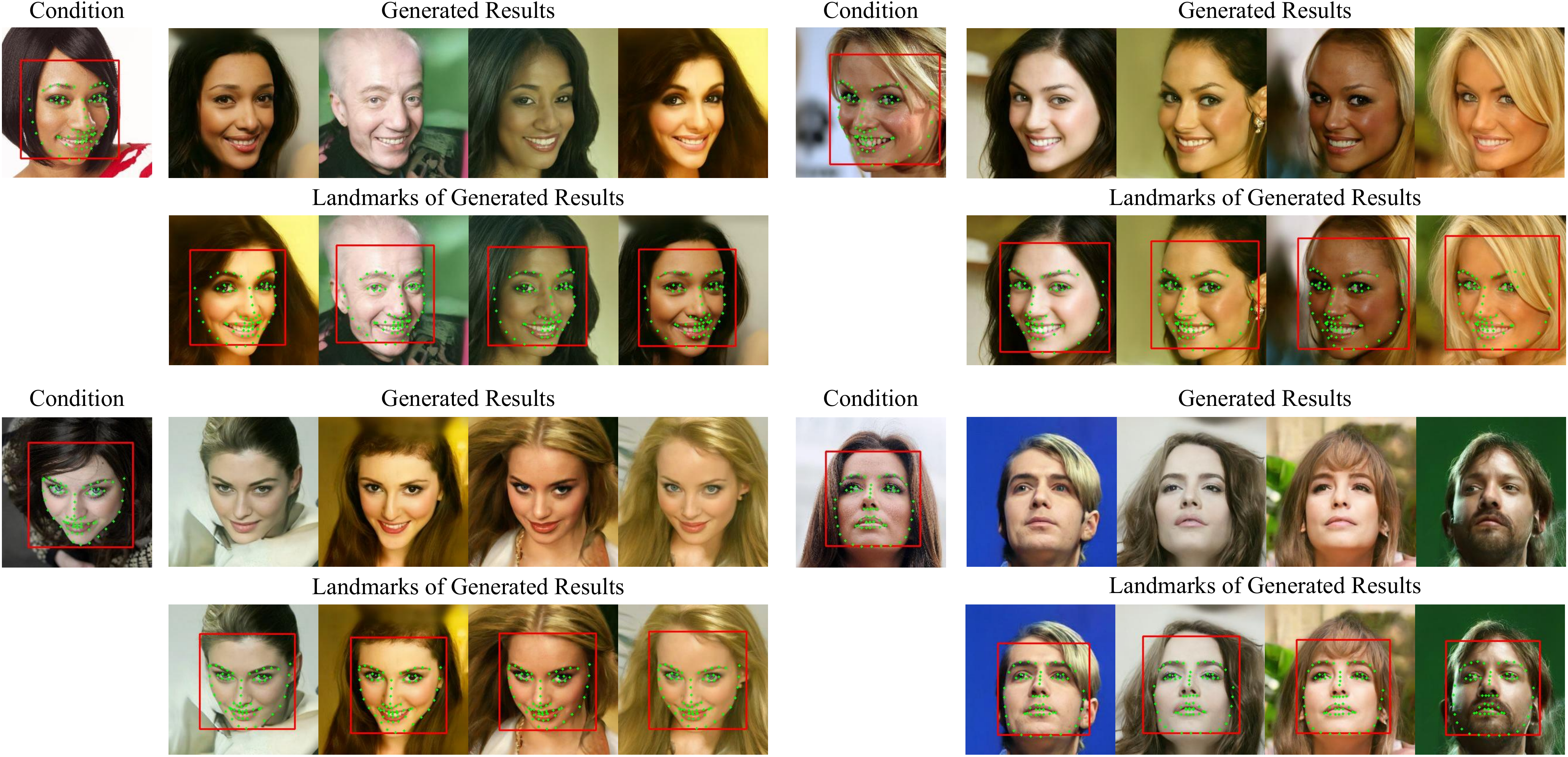}
  \caption{Generated human faces for the landmark-to-image task. We selected landmarks of four faces from different angles to guide the generation process and output the landmarks of the generated results to check the matching degree with given conditions. These results are consistent with the given conditions and have good diversity.}
\label{fig:land2face} 
\end{figure*}

\begin{figure*}[ht]
  \centering
\includegraphics[width=1\linewidth]{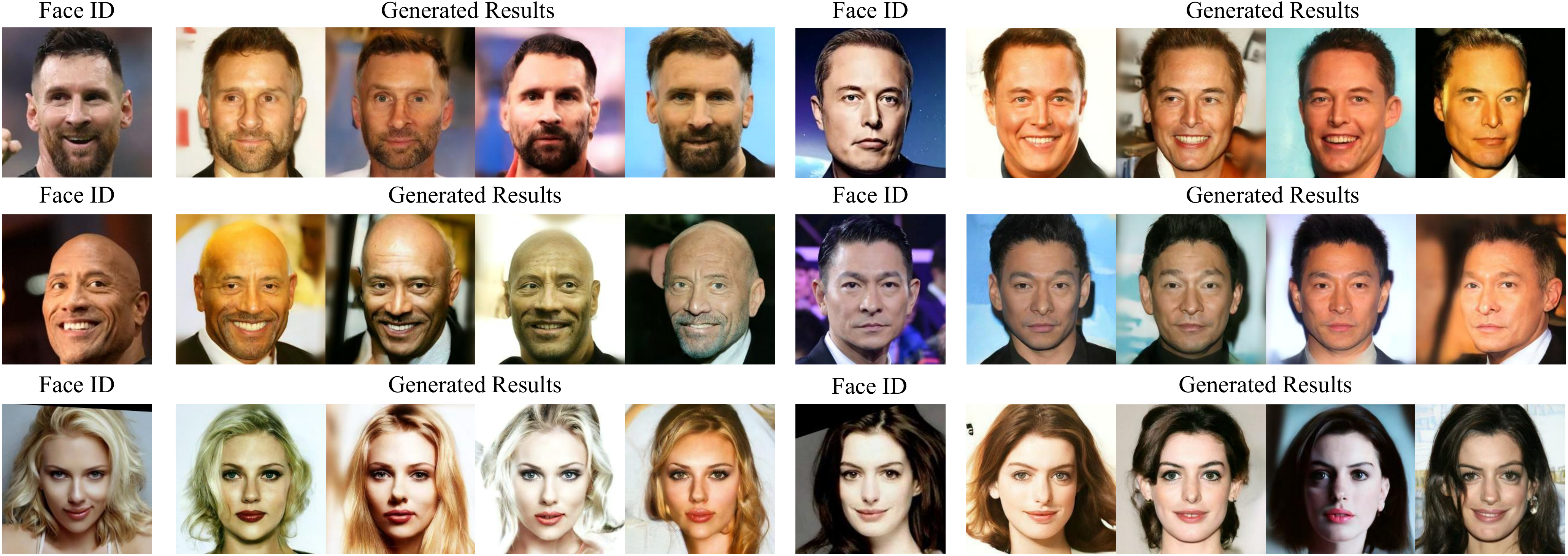}
  \caption{Generated human faces for ID-to-image task. We choose the face IDs of six celebrities as the reference to guide the generation process. These results are consistent with the given conditions and have good diversity.}
\label{fig:id2face} 
\end{figure*}

\begin{figure*}[ht]
  \centering
\includegraphics[width=1\linewidth]{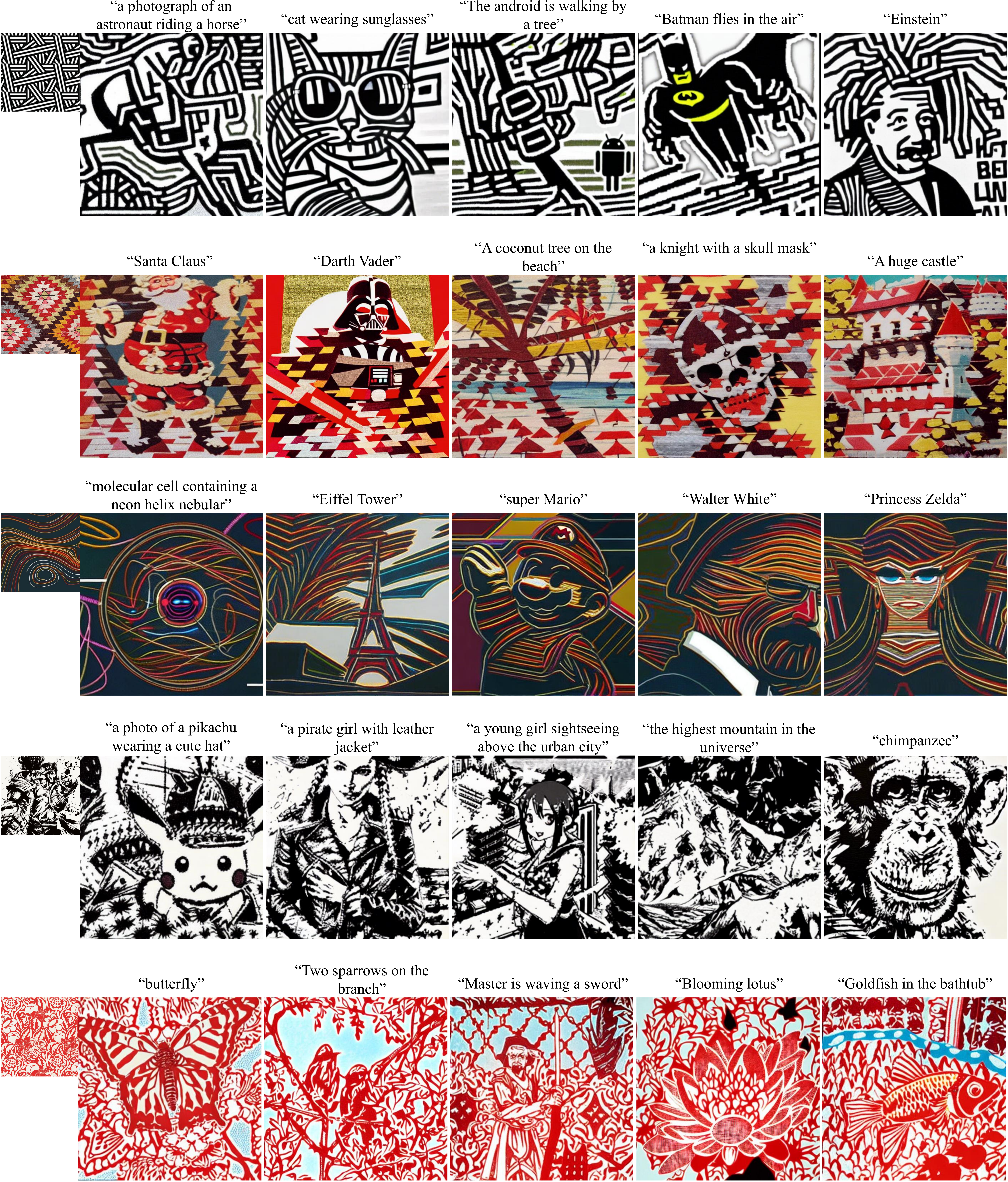}
  \caption{Generation results of training-free style guidance with text-to-image Stable Diffusion~\cite{rombach2022high}. We choose five style images to guide the style of the results generated by Stable Diffusion. These generated results well match the provided style. \textbf{Zoom in for best view.}}
\label{fig:sd_style} 
\end{figure*}

\begin{figure*}[ht]
  \centering
\includegraphics[width=1\linewidth]{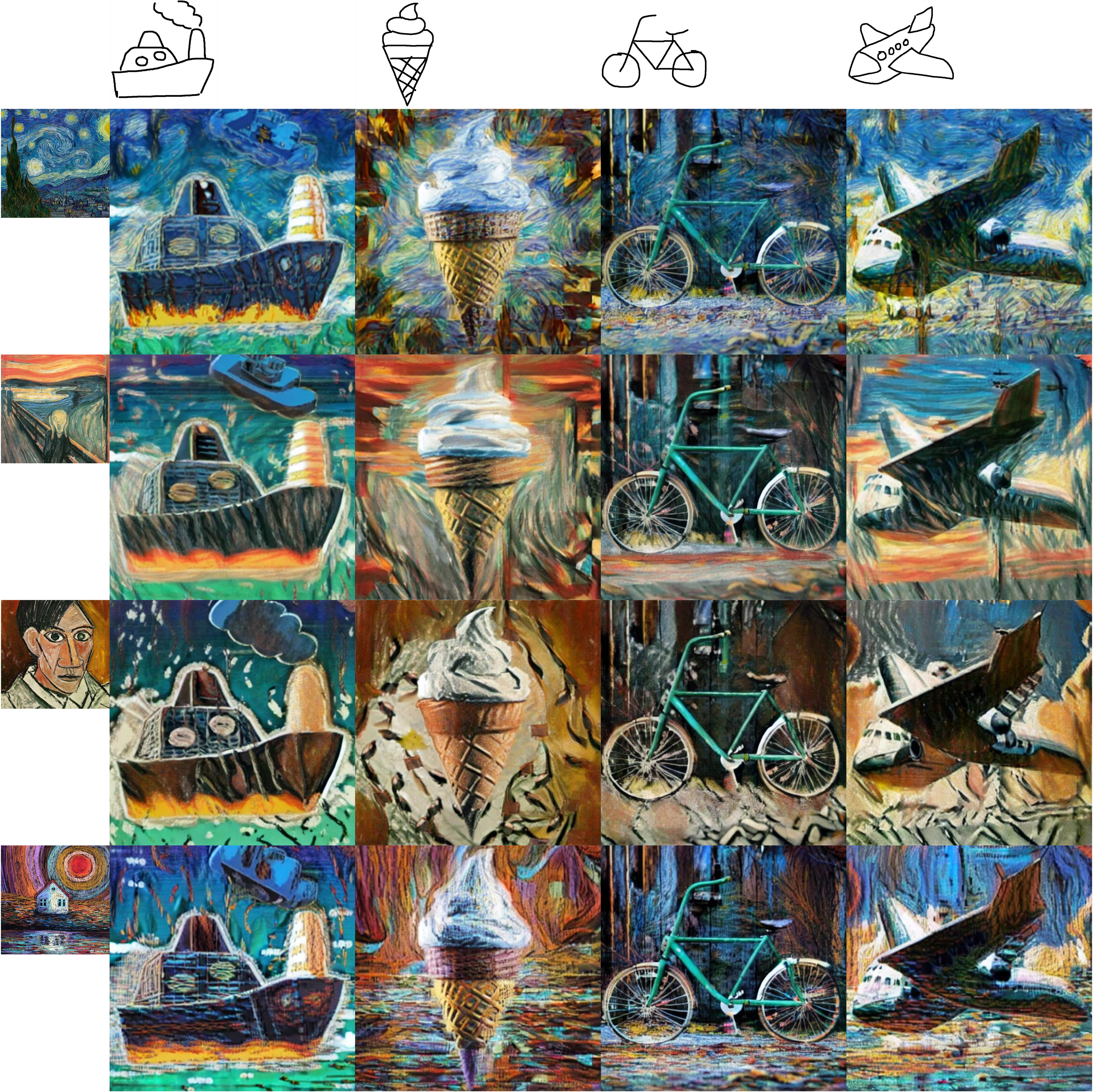}
  \caption{Generated results of training-free style guidance with Scribble ControlNet~\cite{zhang2023adding}. We choose four style images to guide the style of results generated by ControlNet. These generated results well match the provided style. \textbf{Zoom in for best view.}}
\label{fig:cn_style} 
\end{figure*}

\begin{figure*}[ht]
  \centering
\includegraphics[width=1\linewidth]{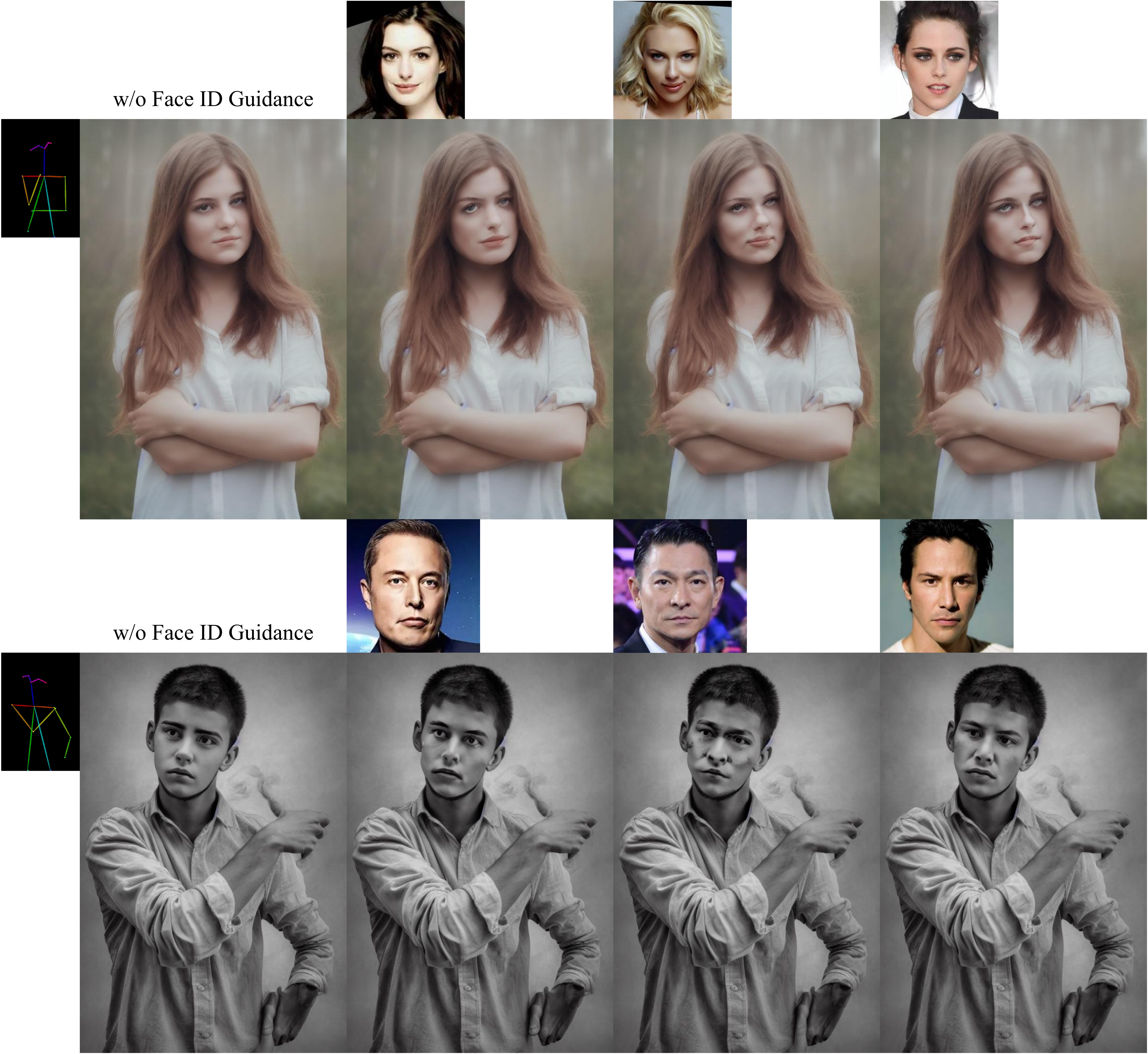}
  \caption{Generated Results of face ID guidance with Human-pose ControlNet~\cite{zhang2023adding}. By fixing random seeds, we can see the effects before and after introducing the ID guidance. These ID-guided results well match the given IDs in the face area. \textbf{Zoom in for best view.}}
\label{fig:cn_id} 
\end{figure*}

\begin{figure*}[ht]
  \centering
\includegraphics[width=1\linewidth]{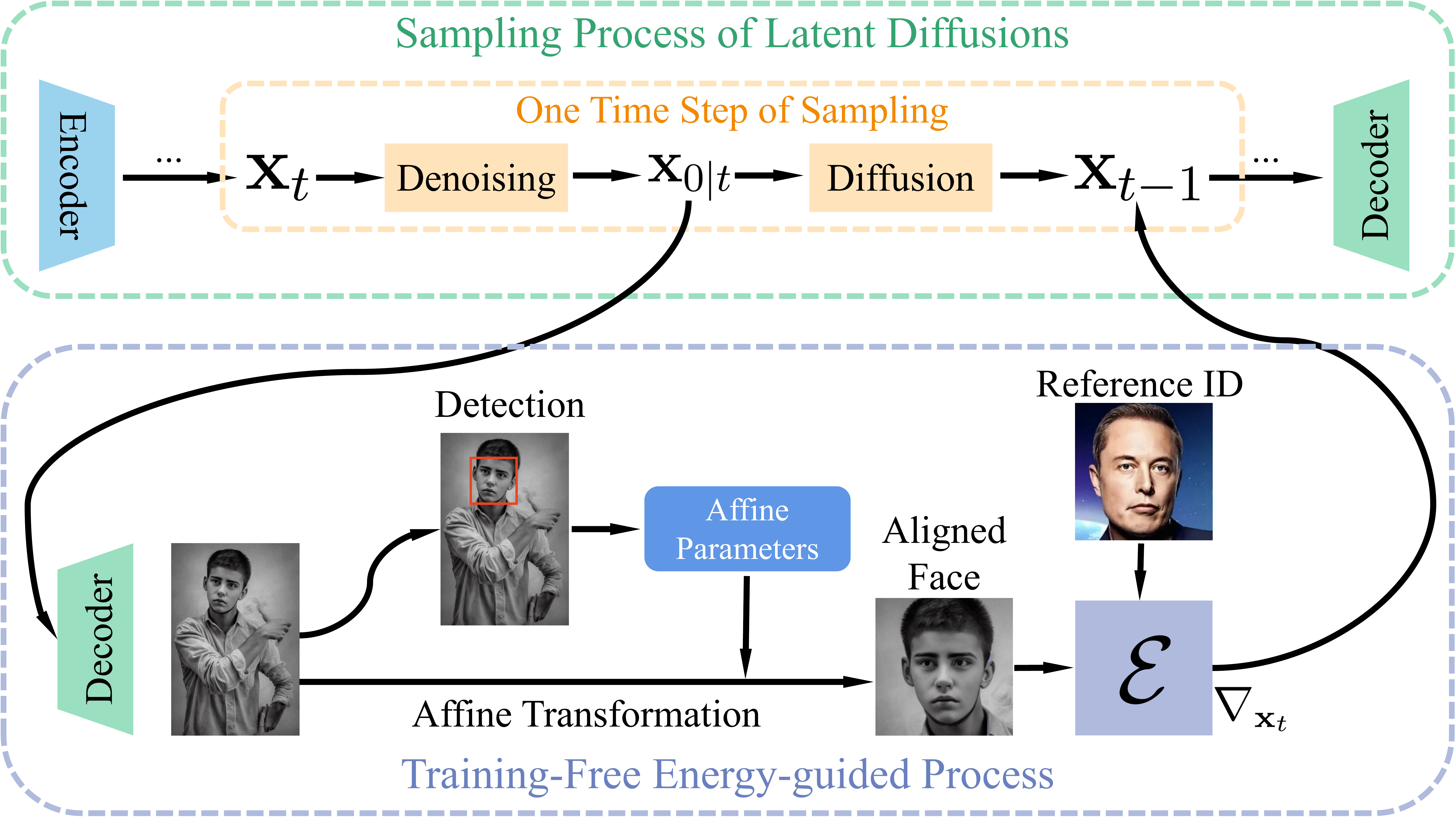}
  \caption{Visualization of the whole training-free face ID guidance process using FreeDoM in Fig.~\ref{fig:cn_id}. We first decode the clean latent code $\mathbf{x}_{0|t}$ into the image domain. Then we detect the position of the human face and the corresponding landmarks. After getting the landmarks, we compute the affine parameters, which are used to perform an affine transformation to extract the aligned face area from the original decoded image. Finally, we compute the ID-based energy function between the aligned and reference faces. The gradient of the energy function to $\mathbf{x}_t$ will be used to update $\mathbf{x}_{t-1}$. Note that the computation of the Decoder and affine transformation is all differentiable, so the energy gradient to $\mathbf{x}_t$ is computable. \textbf{Zoom in for best view.}}
\label{fig:cn_id_process} 
\end{figure*}
\clearpage

\section{Relationship between FreeDoM and Zero-Shot Image Restoration Methods}
\label{sec:zir}

The proposed FreeDoM is a framework that can support various conditions, including the degraded images in the image restoration tasks. Many existing zero-shot image restoration methods~\cite{Choi_2021_ICCV, chung2023diffusion, Chung_2022_CVPR, chung2022improving, kawar2022denoising, Lugmayr_2022_CVPR, song2023pseudoinverseguided, song2021solving, wang2023zeroshot,
wang2023unlimited} can be considered special cases of FreeDoM. Their idea can be summarized as updating the clean intermediate result $\mathbf{x}_{0|t}$ to meet the data consistency constraint, $\mathbf{y}=\mathcal{A}(\mathbf{x}_{0|t})$, where $\mathbf{y}$ is a degraded image and $\mathcal{A}(\cdot)$ is a linear or non-linear degradation operator. When dealing with linear degradation, the degradation operator $\mathcal{A}(\cdot)$ can be written into a matrix $\mathbf{A}$.

Since the image restoration tasks can also be seen as particular conditional generation tasks, these zero-shot image restoration methods can also be explained using the framework of FreeDoM. Take two typical examples: DPS~\cite{chung2023diffusion} uses $-\nabla_{\mathbf{x}_t}||\mathbf{y}-\mathcal{A}(\mathbf{x}_{0|t})||_2^2$ to update the intermediate results, which can be interpreted as a distance measurement function without learning parameters to improve the matching degree between the restored image $\mathbf{x}_{0|t}$ and the degraded image $\mathbf{y}$ in the measurement space; DDNM~\cite{wang2023zeroshot} obtains that the update direction for linear noiseless tasks is $-\mathbf{A}^{\dagger}(\mathbf{A}\mathbf{x}_{0|t}-\mathbf{y})$ through the derivation of Range-Null Space Decomposition, which can also be interpreted as an approximated analytical solution of the gradient of the distance measurement function in DPS on linear cases.


\end{appendices}
}

\end{document}